%% file: main.tex
\documentclass[11pt]{article}
\usepackage[]{latex/acl} 
\usepackage{times}
\usepackage{latexsym}
\usepackage[T1]{fontenc}
\usepackage[utf8]{inputenc} 
\usepackage{microtype}
\usepackage{inconsolata}
\usepackage{graphicx}
\usepackage{booktabs}
\usepackage{multicol,multirow}
\usepackage{array, makecell}
\usepackage{enumitem}
\setitemize{noitemsep}
\usepackage{amssymb}
\usepackage{amsmath}
\usepackage{pifont}
\usepackage{cancel}
\usepackage[edges]{forest}
\usepackage{xspace}
\usepackage{rotating}
\usepackage{adjustbox}
\usepackage{diagbox}
\usepackage{float}

\title{MentalManip: A Dataset For Fine-grained Analysis of Mental Manipulation in Conversations}

\author{
Yuxin Wang$^1$ \quad 
Ivory Yang$^2$ \quad 
Saeed Hassanpour$^3$ \quad 
Soroush Vosoughi$^4$ \\
$^{1,2,4}${Department of Computer Science, Dartmouth College} \\
$^3${Department of Biomedical Data Science, Dartmouth College} \\
$^1${\texttt{Yuxin.Wang.GR@dartmouth.edu}}\\
$^4${\texttt{Soroush.Vosoughi@dartmouth.edu}}
}

\newcommand{\datasetname}{$\textsc{MentalManip}$\xspace}
\newcommand{\datasetnamecon}{$\textsc{MentalManip}_{\text{con}}$\xspace}
\newcommand{\datasetnamemaj}{$\textsc{MentalManip}_{\text{maj}}$\xspace}
\newcommand{\highlightcolor}{black}

\begin{document}
\maketitle
\begin{abstract}

Mental manipulation, a significant form of abuse in interpersonal conversations, presents a challenge to identify due to its context-dependent and often subtle nature. The detection of manipulative language is essential for protecting potential victims, yet the field of Natural Language Processing (NLP) currently faces a scarcity of resources and research on this topic. Our study addresses this gap by introducing a new dataset, named \datasetname, which consists of $4,000$ annotated movie dialogues. This dataset enables a comprehensive analysis of mental manipulation, pinpointing both the techniques utilized for manipulation and the vulnerabilities targeted in victims. Our research further explores the effectiveness of leading-edge models in recognizing manipulative dialogue and its components through a series of experiments with various configurations. The results demonstrate that these models inadequately identify and categorize manipulative content. Attempts to improve their performance by fine-tuning with existing datasets on mental health and toxicity have not overcome these limitations. We anticipate that \datasetname will stimulate further research, leading to progress in both understanding and mitigating the impact of mental manipulation in conversations.
\end{abstract}

\input{01_Introduction}
\input{02_Related_Work}
\input{03_Construction}
\input{04_Experiment}
\input{05_Conclusion}
\input{06_Limitation}
\input{07_Ethics}
\input{08_Acknowledgment}

\bibliography{citation}

\newpage
\appendix
\input{appendix}

\end{document}

%% file: 01_Introduction.tex
\input{tables/data_stats}

\section{Introduction}
\begin{figure}[t]
  \centering
  \includegraphics[width=\linewidth]{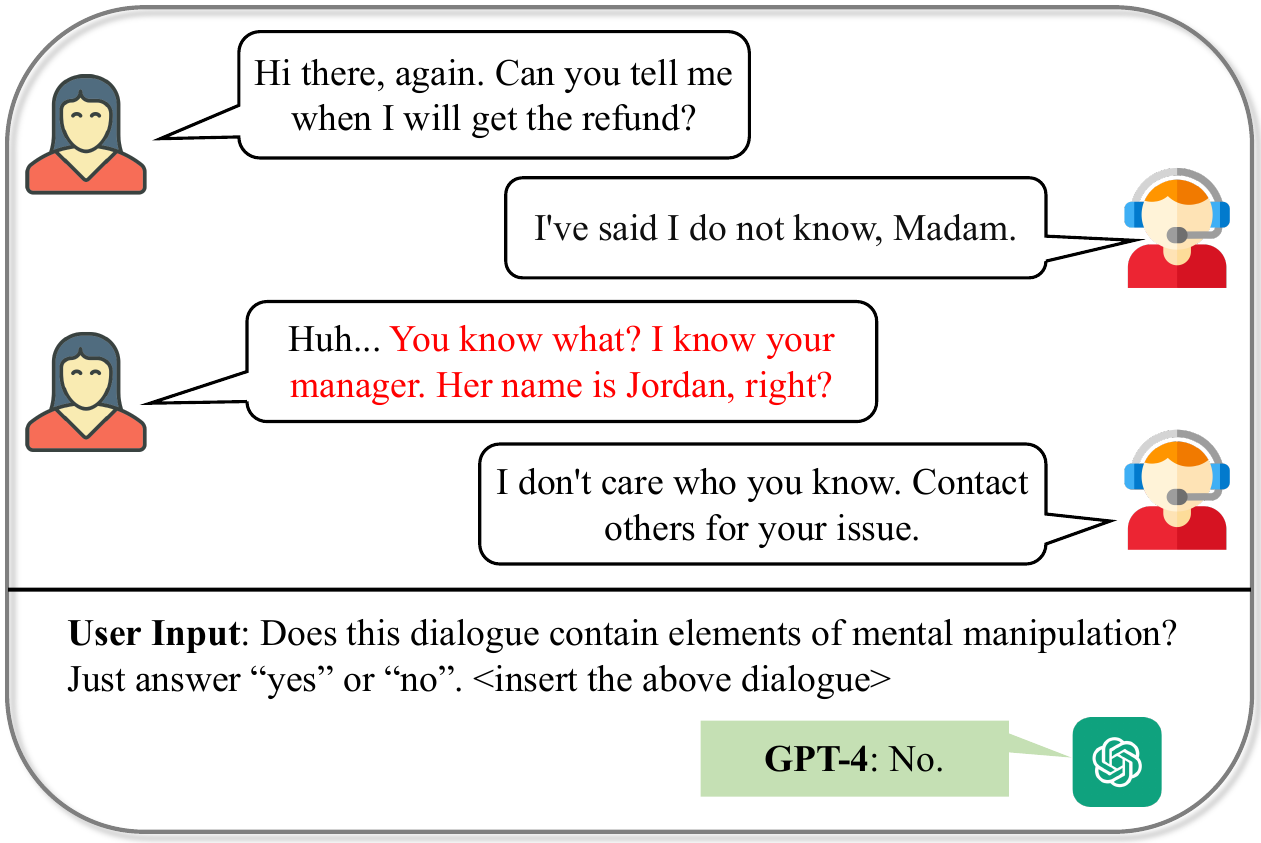}
  \vspace{-5mm}
  \caption{An example dialogue that contains elements of mental manipulation which GPT-4 fails to identify ($Temperature = 0$). The manipulative parts are highlighted in red.}
  \vspace{-4mm}
  \label{fig:illustration}
\end{figure}

Mental manipulation is a deceptive strategy aimed at controlling or altering someone's thoughts and feelings to serve personal objectives~\cite{barnhillManipulation2014}. 
Facilitated by digital technologies, mental manipulation has gained unprecedented capability to target and influence individuals via interpersonal interactions and public dissemination of information ~\cite{iencaAIManipulation2023}, causing significant mental health distress to victims~\cite{hamelAbuse2023}.
Compared to overt verbal toxicity and abuse, such as hate speech, manipulation is more insidious, deceitful, and context-dependent, posing challenges for individuals and automatic moderation tools to discern.
For years, the NLP community has witnessed advancements in verbal toxicity and abuse detection, but most of those works focus on context-free content and face challenges in identifying implicit toxicity~\cite{wiegandImplicitlyAbusiveLanguage2021,mishraTacklingOnlineAbuse2020,yinHiddenObviousMisleading2022,dengRecentAdvancesSafe2023}.
Existing works in dialogue systems have targeted context-aware toxicity, but are limited to direct toxicity, such as profanity, condescension and forms of hate speech~\cite{bahetiJustSayNo2021,gaoDetectingOnlineHate2017,wangTalkDownCorpusCondescension2019}. 
We argue that existing toxicity detection resources are insufficient for developing automatic systems to detect and properly handle verbal mental manipulation. 
Additionally, current state-of-the-art Large Language Models (LLMs) are not well-positioned to address this issue, as demonstrated in Figure~\ref{fig:illustration}. 

This paper introduces \datasetname, a dataset with multi-level annotations for mental manipulation detection and classification. 
We define mental manipulation as \textit{using language to influence, alter, or control an individual's psychological state or perception for the manipulator's benefit}. 
\textcolor{\highlightcolor}{This definition aligns with one of the explanations of manipulative influence in Barnhill's work on philosophy of online manipulation~\cite{barnhill2022philosophy}.}
\datasetname dataset contains $4,000$ multi-turn fictional dialogues between two characters extracted from online movie scripts. 
To enable fine-grained analysis, we proposed a labeling taxonomy covering three dimensions: presence of manipulation, manipulation technique, and targeted vulnerability, which are illustrated in Figure~\ref{tree:taxonomy_tree}. 
Our taxonomy aids in the precise detection of mental manipulation and facilitates a nuanced classification of the techniques used by manipulators, as well as the vulnerabilities targeted in victims.

We conducted three classification tasks on \datasetname to detect the existence of mental manipulation and its elements.
Our experiments spanned state-of-the-art generative and discriminative LLMs across multiple settings.
To investigate the effectiveness of existing datasets in toxic speech and mental health for our objectives, we fine-tuned LLMs with seven relevant datasets and conducted evaluations.
Experimental results reveal that these LLMs are limited in understanding mental manipulation, as shown by a significant number of misclassified ``manipulative'' dialogues.
Moreover, fine-tuning LLMs with these relevant datasets did not enhance their detection and classification capabilities on manipulative language.
These findings highlight the importance of our dataset, and suggest an avenue for future studies in mental manipulation detection and analysis.

\textcolor{\highlightcolor}{Our \datasetname dataset and the code for statistics analysis and experiments in this paper are available in our GitHub repository: \url{https://github.com/audreycs/MentalManip}.}

%% file: tables/data_stats.tex
{
\renewcommand{\arraystretch}{0.8}
\begin{table*}[t]
    \setlength{\tabcolsep}{3pt}
    \centering
    \small
    \resizebox{\textwidth}{!}{%
    \begin{tabular}{ccccccc}
    \toprule
    \textbf{Dataset} & \textbf{Research Scope} & \textbf{Dialogical} & \textbf{\#Sample} & \textbf{\#Avg. Utterance} & \textbf{Data Source} & \textbf{Label Scheme}
    \\ \toprule
    Dreaddit~\cite{turcanDreadditRedditDataset2019} & Mental Stress & No & $~~~~3,553$ & $1$ & Reddit posts & Binary
    \\ \midrule
    SDCNL~\cite{haqueDeepLearningSuicide2021} & Depression \& Suicide & No & $~~~~1,895$ & $1$ & Reddit posts & Binary
    \\ \midrule
    ToxiGen~\cite{hartvigsenToxiGen2022} & Hate Speech & No & $274,186$ & $1$ & GPT-3 & Binary
    \\ \midrule
    DetexD~\cite{yavnyiDeTexDBenchmarkDataset2023} & Delicate Text & No & $~~~~1,023$ & $1$ & Online forums & Multi-level
    \\ \midrule
    Fox News~\cite{gaoDetectingOnlineHate2017} & Hate Speech & Yes & $~~~~~~~~814$ & $2.0$ $(\pm0.1)$ & News comments & Binary
    \\ \midrule
    TalkDown~\cite{wangTalkDownCorpusCondescension2019} & Condescension & Yes & $~~~~4,992$ & $2$ & Reddit comments & Binary
    \\ \midrule
    ToxiChat~\cite{bahetiJustSayNo2021} & Offensiveness & Yes & $~~~~2,000$ & $2.3$ $(\pm0.5)$  & Reddit comments & Binary
    \\ \midrule
    MDRDC~\cite{zhangTaxonomyDataSet2021} & Malevolence & Yes & $~~~~6,000$ & $4.8$ $(\pm1.9)$ & Twitter threads & Multi-level
    \\ \midrule
    \datasetname~(ours) & Mental Manipulation & Yes & $~~~~4,000$ & $6.5$ $(\pm5.3)$ & Movie scripts & Multi-level
    \\ \bottomrule
    \end{tabular}}
    \vspace{-1mm}
    \caption{Comparison of properties and statistics: \datasetname dataset versus existing datasets in verbal toxicity and mental health problem detection. The columns ``\#Sample'' and ``\#Avg. Utterance'' represent the number of instances/dialogues and the average number of utterances per dialogue, respectively. Numbers in round brackets are standard deviations.}
    \vspace{-3mm}
    \label{tab:stats_compare}
\end{table*}
}

%% file: 02_Related_Work.tex
\section{Related Works}
\subsection{Mental Health Detection}
Leveraging NLP technologies for the early detection and intervention of mental health issues stands as a valuable endeavor.
In recent decades, there has been considerable research identifying specific mental health concerns, such as stress~\cite{guntukuUnderstandingMeasuringPsychological2019,nijhawanStressDetectionUsing2022}, depression~\cite{eichstaedtFacebookLanguagePredicts2018,xuLeveragingRoutineBehavior2019}, and suicide~\cite{dechoudhuryDiscoveringShiftsSuicidal2016,coppersmithNaturalLanguageProcessing2018}.
Scalable and accessible data resources for these issues have been proposed.~\cite{turcanDreadditRedditDataset2019,haqueDeepLearningSuicide2021,naseemEarlyIdentificationDepression2022}. 
More recently, research has shown that LLMs, like GPT-4, exhibit promising yet limited performance on tasks related to mental health~\cite{yangInterpretableMentalHealth2023,xuMentalLLMLeveragingLarge2023}.
Researchers have pointed out a lack of explainability for the detection results of LLMs, and highlighted the importance of domain-specific fine-tuning on LLMs for better performance.
These findings underscore the need for data resources featuring nuanced annotations and targeting unaddressed mental health issues.

\subsection{Toxic Speech Detection}
In NLP, ``toxic speech'' is an umbrella term referring to language that is rude, disrespectful, or offensive, potentially harming conversation quality and negatively impacting recipients~\cite{dixonMeasuringMitigatingUnintended2018}. 
Lots of benchmark datasets have been developed to detect explicit and implicit toxic speech in online posts and comments, including those focusing on racism and sexism~\cite{waseemAreYouRacist2016,basileSemEval2019TaskMultilingual2019,hartvigsenToxiGen2022,yavnyiDeTexDBenchmarkDataset2023}, 
online harassment~\cite{hosseinmardiDetectionCyberbullyingIncidents2015,rosaUsingFuzzyFingerprints2018}, 
and trolling~\cite{miaoDetectingTrollTweets2020}.
Recent works have also investigated performance of state-of-the-art LLMs on toxic speech ~\cite{wangDecodingTrustComprehensiveAssessment2023}.
Although many mental manipulations, such as intimidation, fall under toxic speech, their subtle and complex nature create challenges beyond the capability of context-free toxicity detection methods.
Existing works in dialogue systems address context-aware toxicity detection~\cite{wangTalkDownCorpusCondescension2019,bahetiJustSayNo2021,zhangTaxonomyDataSet2021}, but they focus on explicit toxicity and overlook implicit verbal manipulation.

Table~\ref{tab:stats_compare} summarizes some existing datasets addressing toxicity or mental health problems.

%% file: 03_Construction.tex
\section{Constructing \datasetname}
\subsection{Taxonomy}
Establishing a structured labeling taxonomy when developing a dataset is crucial.
Drawing inspiration from Simon's research on psychological manipulation~\cite{simon2011sheep}, we crafted a multi-level taxonomy encompassing three dimensions:
\begin{itemize}[noitemsep,topsep=2pt,parsep=2pt,partopsep=2pt]
    \item Presence of Manipulation: This level employs binary classification, indicating if a dialogue contains elements of mental manipulation.
    \item Manipulation Technique: This level identifies specific manipulation techniques used in conversation.
    \item Targeted Vulnerability: The last level indicates particular victim vulnerabilities exploited by the manipulator.
\end{itemize}
We present the detailed taxonomy in Figure~\ref{tree:taxonomy_tree}, which contains $11$ different techniques and $5$ vulnerabilities.
We expatiate the definition of each technique and vulnerability in Appendix~\ref{appendix:definition}.
To ensure clarity and comprehensiveness, we incorporated insights from a psychological expert and feedback from annotators.

\input{tables/taxonomy_tree}

\subsection{Data Source and Preprocessing}
We prioritize dialogues as our primary data format as they maintain original context, unlike standalone comments and posts. 
To guarantee a semantically rich and stylistically diverse dataset, we prioritize human-crafted content over LLM-generated material.
We finally chose Cornell Movie Dialogs Corpus\footnote{\url{https://www.cs.cornell.edu/~cristian/Cornell_Movie-Dialogs_Corpus.html}}~\cite{danescu-niculescu-mizilChameleonsImaginedConversations2011} as the data source to construct \datasetname. 
The Cornell Movie Dialogs Corpus contains $220,579$ conversational exchanges extracted from $617$ raw movie scripts spanning a wide range of genres. 
The overwhelming majority of dialogues occur between two characters, which we standardized for.
We replaced original speakers' names with ``Person1'' and ``Person2'' to eliminate potential biases.

Since manipulative language is relatively sparse in conversation, we need to filter the original data to get dialogues potentially containing elements of manipulation.
We utilized two approaches to achieve this: 1) key phrase-based matching, and 2) BERT classification.
\textcolor{\highlightcolor}{
For key phrase-based matching, we sourced key phrases from online resources, selecting those that frequently occur in manipulative conversations, without restricting their n-gram size. 
After collection, we manually conducted tense conversion (convert all phrases to present tense), phrase simplification (e.g., ``It’s fine, nobody cares about me anyway'' to ``nobody cares about me'') and merging of similar phrases. 
Ultimately, we obtained a list of $175$ cleaned key phrases.
Appendix~\ref{appendix:key_phrase} presents examples of the cleaned key phrases and details of the online resources we used. 
The full list of cleaned key phrases is available in our GitHub repository.
To screen out candidate dialogues where key phrases are present, we implemented a length-adaptive matching criterion due to the lexical diversity of language. 
A dialogue is considered a match if any sentence contains at least $P\%$ tokens from a key phrase. 
The value of $P$ is detailed in Table~\ref{tab:phrase_matching}.
}

For BERT classification approach, we fine-tuned a pre-trained BERT model with a sequence classification head on top.
Our goal was to get a classifier to differentiate manipulative dialogues from general toxic content.
To prepare the training data, we inquired GPT-4 Turbo~\cite{bubeckSparksArtificialGeneral2023} by zero-shot prompting\footnote{API calling format is presented in Appendix~\ref{appendix:prompt}.} on whole Cornell Movie Dialogs Corpus and obtained a set of ``manipulative'' dialogues flagged by GPT-4.
We observed that GPT-4 generated a large portion of false positives for manipulative content.
We examined $1,378$ ``manipulative'' dialogues flagged by GPT-4, and labeled only $464$ dialogues as truly manipulative, with the remaining 914 being false positives.
These $1,378$ labeled dialogues were then used to train the BERT classifier.
Finally, we employed BERT classifier on all identified ``manipulative'' dialogues to obtain highly likely manipulative dialogues.

\input{tables/phrase_matching}

\textcolor{\highlightcolor}{In total, we obtained $1,406$ dialogues from key phrase-based matching and $3,739$ dialogues from BERT classification.
We then removed duplicates and low-quality dialogues, such as extremely short dialogues with broken contexts, and rephrased some samples to enhance readability. 
The final count of dialogues for annotation was $4,876$.}

\subsection{Human Annotation} \label{subsec:annotation}
We established our annotation platform using Label Studio\footnote{\url{https://labelstud.io/}}. 
Each dialogue represents an annotation task. 
\textcolor{\highlightcolor}{
We recruited $17$ college students who are native or fluent English speakers as annotators. 
The annotator pool reflects diversity across gender (14 females, 3 males), ethnicity (5 Whites, 11 Asian, 1 Latino), educational backgrounds (e.g., English, Computer Science, Physics majors), and cultural backgrounds (e.g., US-born and non-US-born).}
During recruitment, applicants with an educational background in psychology or linguistics were preferred. 
We conducted tutorial sessions, required annotators to carefully read instructions, and monitored their annotation activities.
Screenshots of the annotation platform and instructions are provided in Appendix~\ref{appendix:screenshot}.
To ensure annotation quality, we assigned three annotators to each task.
\textcolor{\highlightcolor}{
During task assignment, we avoided assigning the same pairs of annotators to evaluate the same dialogues, thereby reducing the potential for bias in assessing inter-annotator agreement.}

In each task, annotators are presented a dialogue, then prompted to answer four questions:
\begin{itemize}[noitemsep,topsep=1pt,parsep=1pt,partopsep=1pt]
    \item $\mathcal{Q}1$ (binary choice): Does this dialogue contain elements of mental manipulation? (Options are ``Yes'' or ``No''.)
    \item $\mathcal{Q}2$ (multiple choice): What techniques are used by the manipulator? (Options are techniques in Figure~\ref{tree:taxonomy_tree}.)
    \item $\mathcal{Q}3$ (binary choice): Are there any victims resulting from manipulation in this dialogue? (Options are ``Yes'' or ``No''.)
    \item $\mathcal{Q}4$ (multiple choice): Which vulnerabilities are targeted in the victim? (Options are vulnerabilities in Figure~\ref{tree:taxonomy_tree}.)
\end{itemize}
$\mathcal{Q}2$ and $\mathcal{Q}3$ are conditional upon $\mathcal{Q}1$, and $\mathcal{Q}4$ is conditional upon $\mathcal{Q}3$.
Annotators could choose at most three techniques and at most two vulnerabilities.
To accommodate indecision, we included a ``cannot decide'' option in $\mathcal{Q}2$ and $\mathcal{Q}4$.
Annotators were required to rate their confidence on a scale from $1$ (not confident) to $5$ (very confident). Furthermore, annotators could highlight sections they identified as manipulative to aid in our review.
Appendix~\ref{appendix:annotation_example} provides an annotation example.

In total, we obtained more than $13$K annotations.
After quality review, the final size of well-labeled dialogues is $4,000$.
Appendix~\ref{appendix:annotation_quality} provides a detailed statistics of annotation quality, including the heat map of agreement between any two annotators, inter-gender agreement, scatter plot of agreement and confidence, and density distributions of agreement and confidence among annotators. 
We also calculated the inter-annotator agreement using Fleiss’ Kappa~\cite{fleissEquivalenceWeightedKappa1973} based on their answers on $\mathcal{Q}1$. 
The score was \textcolor{\highlightcolor}{$0.596$}, indicating a moderate annotator agreement. 
This agreement level is as per our expectation, as the judgment of manipulation is very subjective.
We name this dataset \datasetname, and provide samples of it as a supplementary file. 

\begin{figure*}[t!]
  \begin{minipage}{0.38\textwidth}
    \centering
    \includegraphics[width=\linewidth]{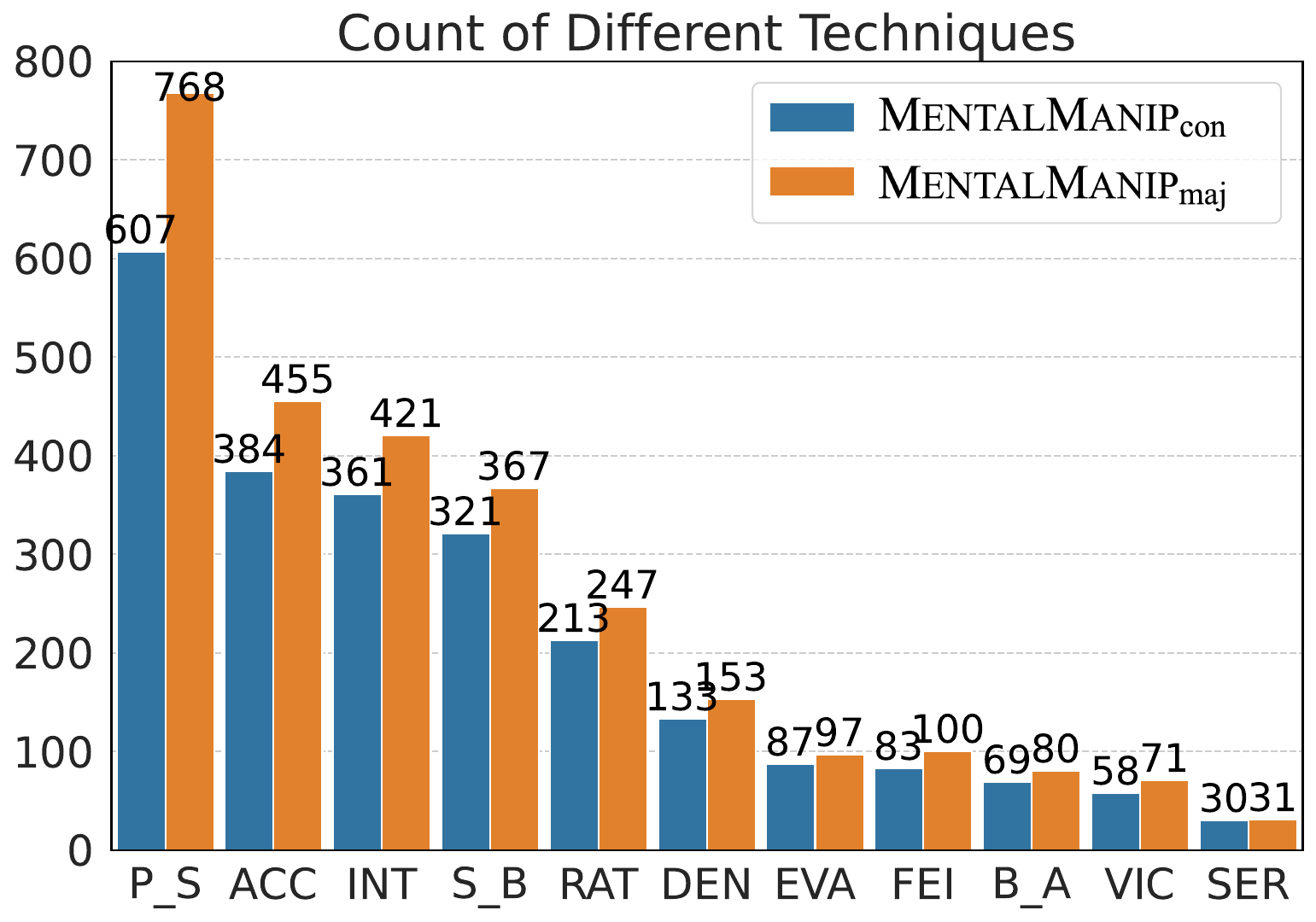}
  \end{minipage}
  \begin{minipage}{0.265\textwidth}
    \centering
    \includegraphics[width=\linewidth]{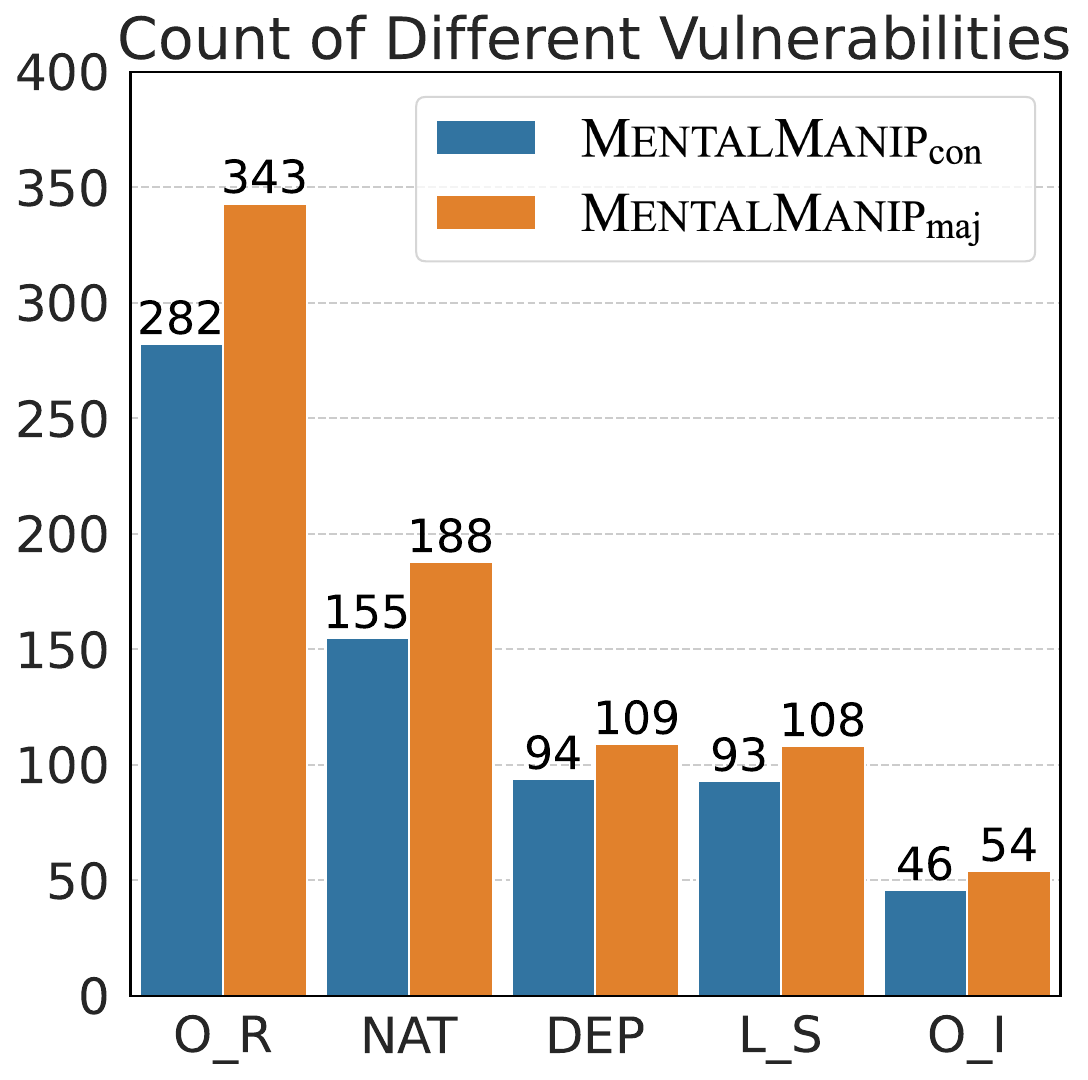}
  \end{minipage}
  \begin{minipage}{0.34\textwidth}
    \centering
    \includegraphics[width=\linewidth]{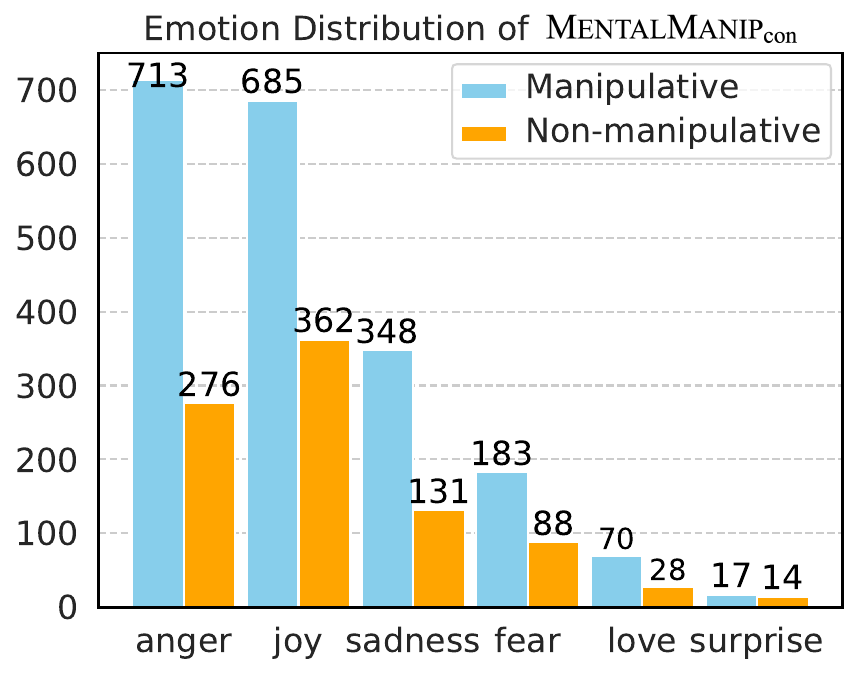}
  \end{minipage}
  \vspace{-2mm}
  \caption{\textcolor{\highlightcolor}{Statistics of \datasetnamecon and \datasetnamemaj. The x-axis ticks in the left two panels are abbreviations for techniques and vulnerabilities (see Appendix~\ref{appendix:definition}). The emotion distribution of \datasetnamemaj dataset is in Appendix~\ref{appendix:statistics_maj}.}}
  \label{fig:mentalmanip_stats1}
\end{figure*}

\begin{figure*}[h!]
  \begin{minipage}{0.32\textwidth}
    \centering
    \includegraphics[width=\linewidth]{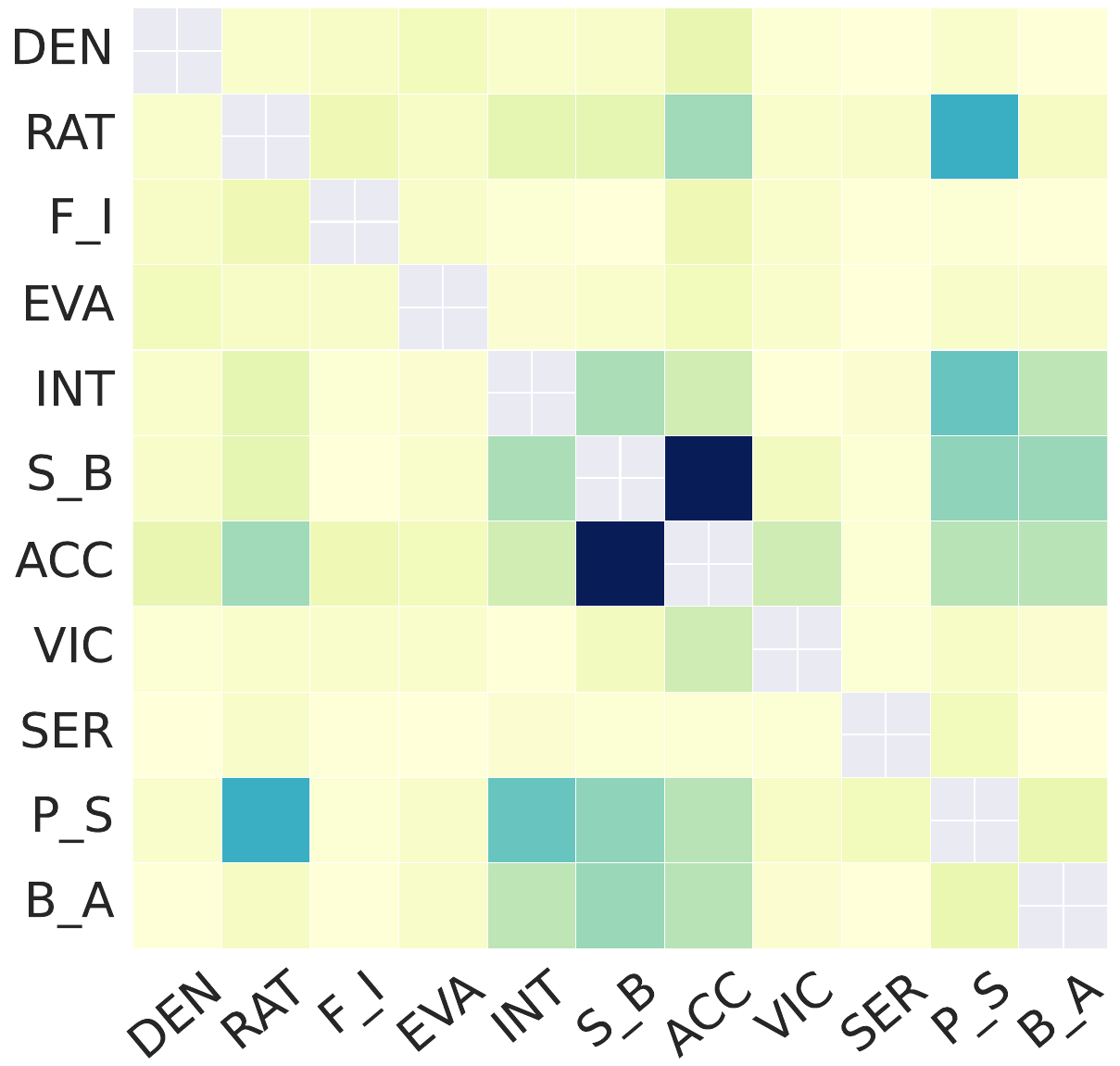}
  \end{minipage}
  \hspace{2pt}
  \begin{minipage}{0.32\textwidth}
    \centering
    \includegraphics[width=\linewidth]{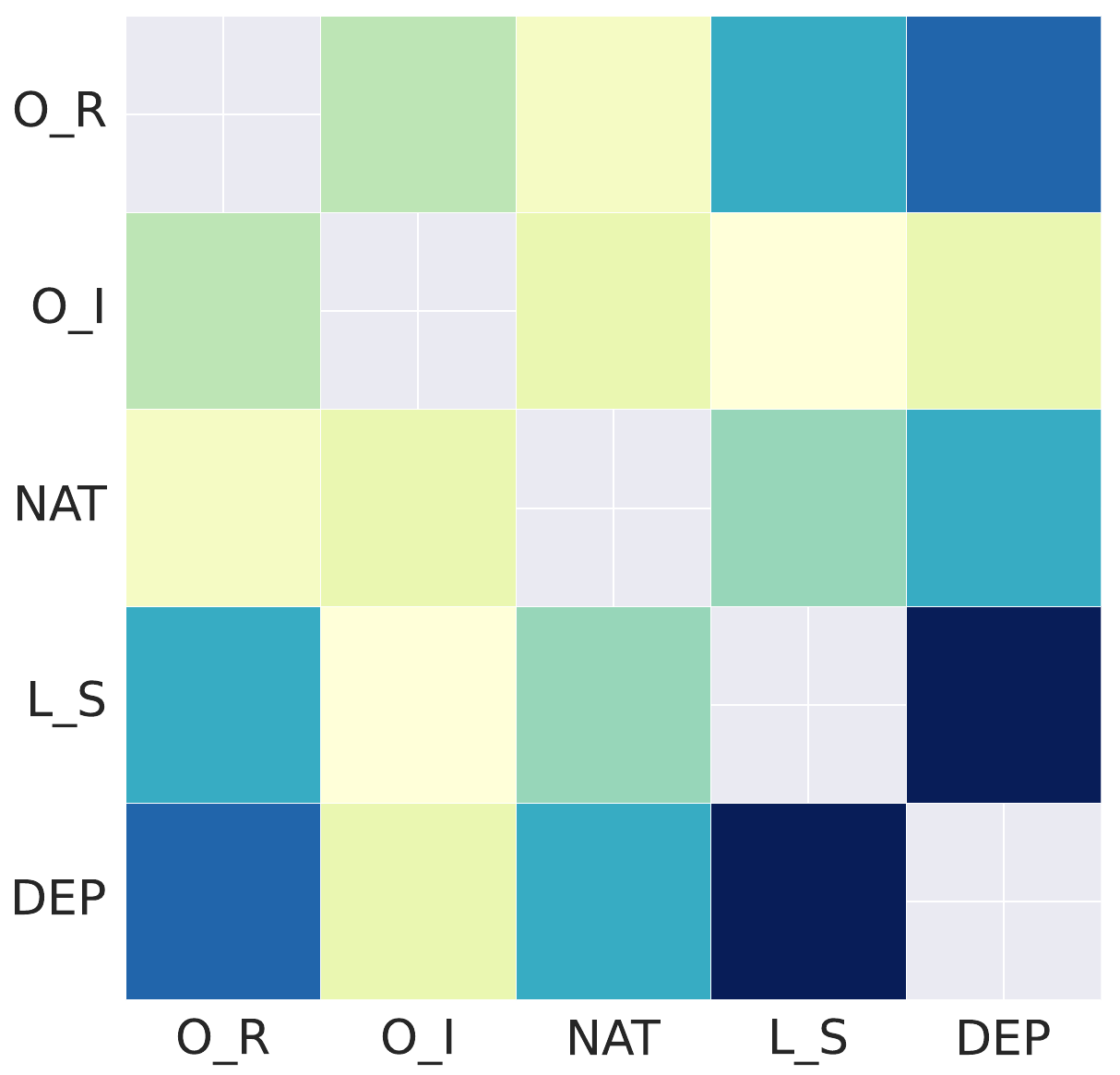}
  \end{minipage}
  \hspace{2pt}
  \begin{minipage}{0.32\textwidth}
    \centering
    \includegraphics[width=\linewidth]{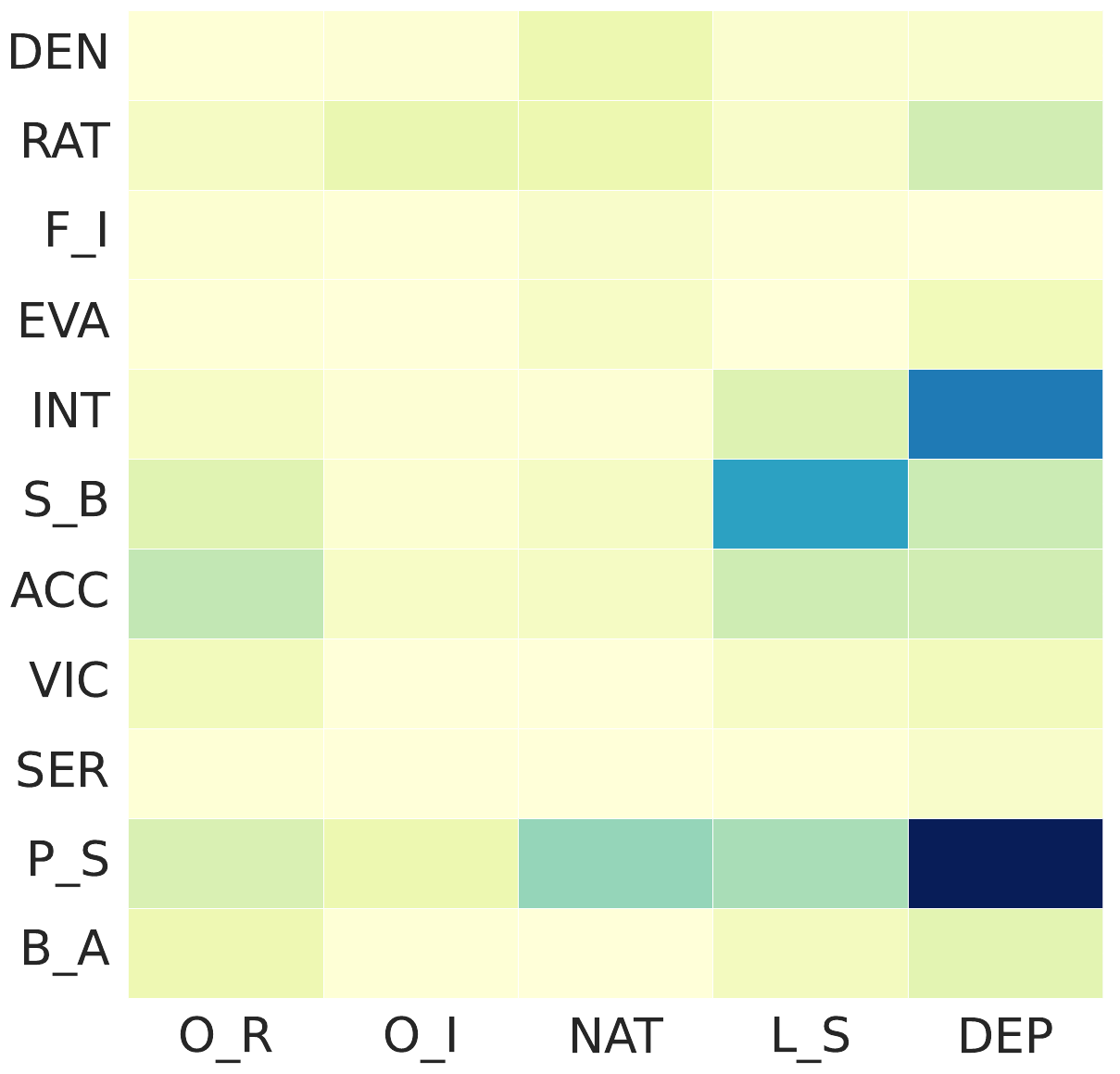}
  \end{minipage}
  \vspace{-1mm}
  \caption{\textcolor{\highlightcolor}{Co-occurrence heat maps among techniques (left), vulnerabilities (center), and techniques and vulnerabilities (right) in \datasetnamecon dataset. Darker cell indicates a higher co-occurrence. The same figures showing results on \datasetnamemaj dataset are in Appendix~\ref{appendix:statistics_maj}.}}
  \label{fig:heat_map}
\end{figure*}

\subsection{Final Label Generation}\label{subsec:label_generation}
To prepare the dataset for experiment, final labels need to be created.
As each dialogue is annotated by three annotators, we adopted two strategies for generating the final labels:
\begin{itemize}[noitemsep,topsep=1pt,parsep=1pt,partopsep=1pt]
    \item Consensus agreement: This strategy only selects dialogues with the same annotation results from all three annotators. The accordant result becomes the final label.
    \item Majority agreement: This strategy adopts the majority rule, where the majority of the annotation results becomes the final label, even if annotators contribute discrepant results.
\end{itemize}

\input{tables/final_label}

\begin{figure*}[h!]
\centering
    \begin{minipage}{0.44\textwidth}
        \centering
        \includegraphics[width=\linewidth]{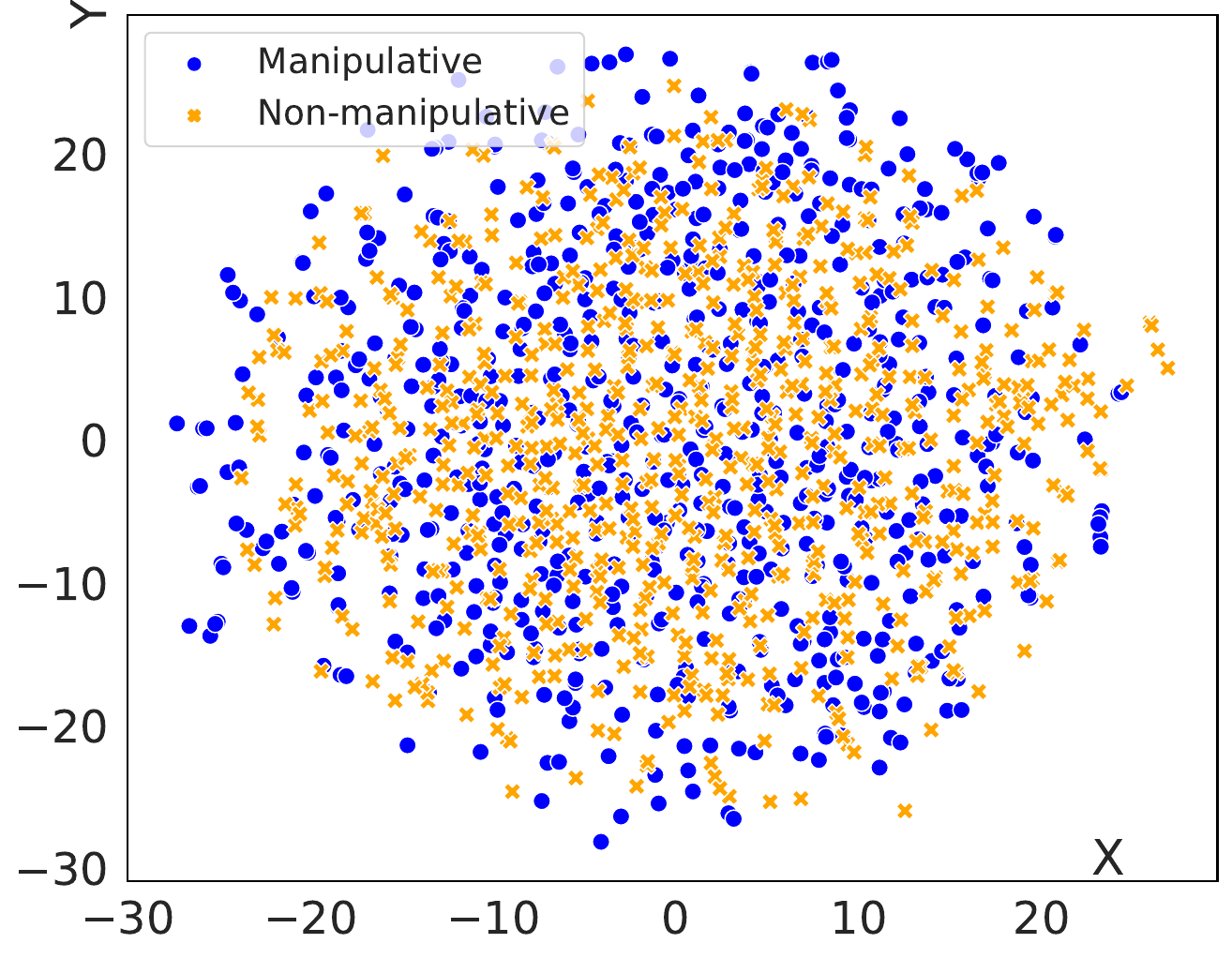}
    \end{minipage}
    \hspace{10pt}
    \begin{minipage}{0.44\textwidth}
        \centering
        \includegraphics[width=\linewidth]{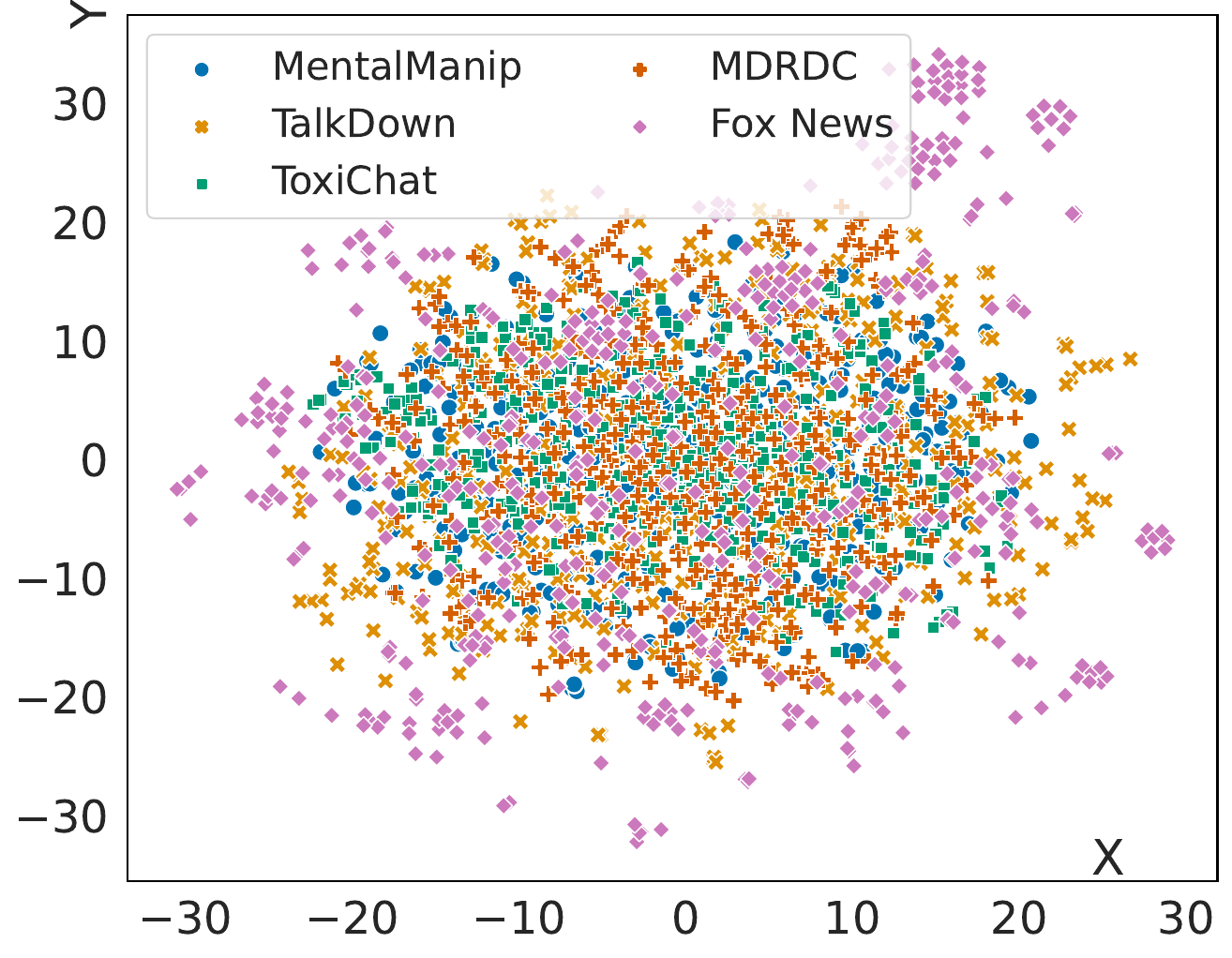}
    \end{minipage}
    \vspace{-1mm}
    \caption{t-SNE visualization of Sentence Transformer embeddings of manipulative and non-manipulative dialogues in \datasetnamecon (left) and the distribution of \datasetname and other dialogical datasets (right).}
    \label{fig:embedding_space}
\end{figure*}

\begin{figure*}[h!]
\centering
    \begin{minipage}{0.45\textwidth}
        \centering
        \includegraphics[width=\linewidth]{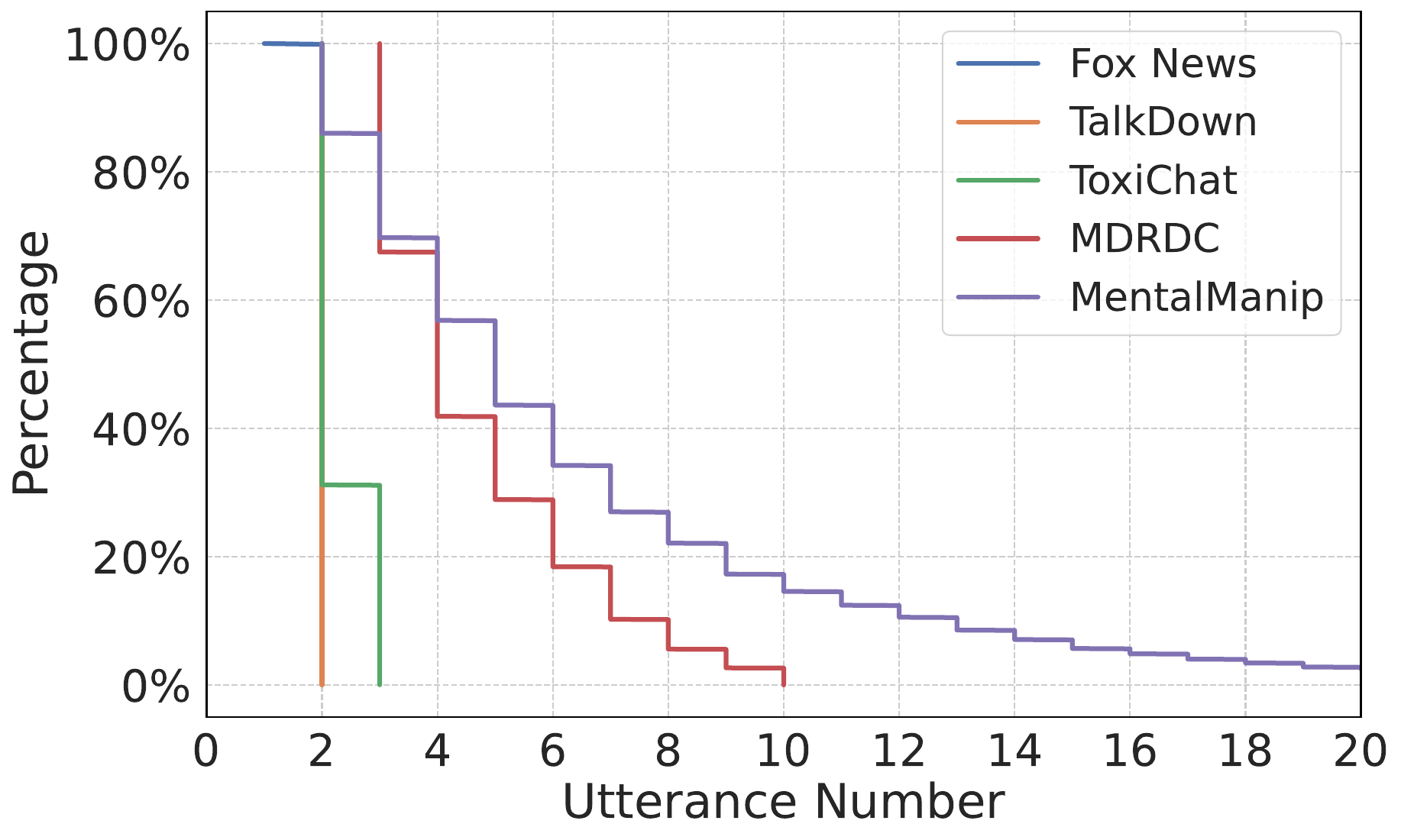}
    \end{minipage}
    \hspace{5pt}
    \begin{minipage}{0.45\textwidth}
        \centering
        \includegraphics[width=\linewidth]{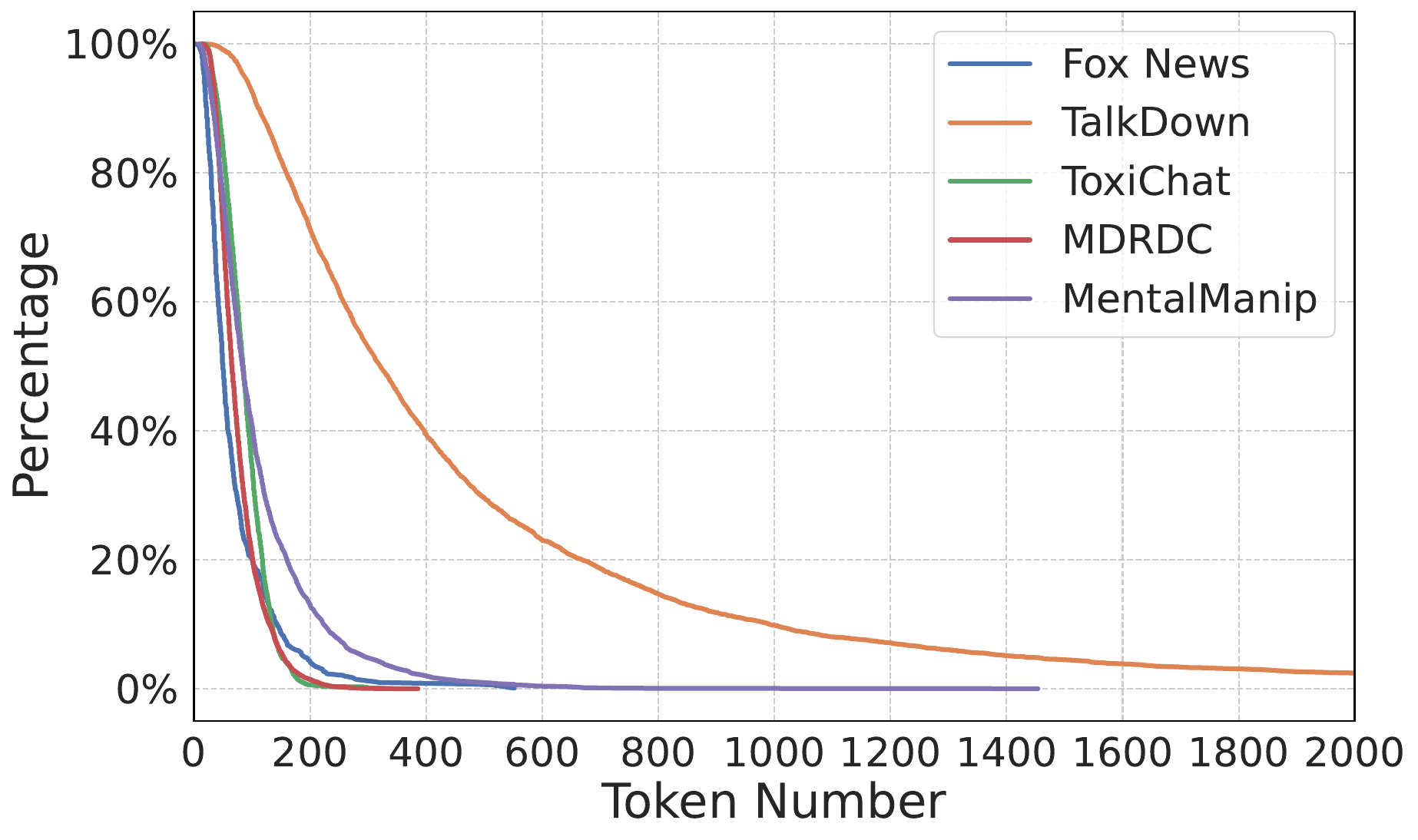}
    \end{minipage}
    \vspace{-2mm}
    \caption{CCDF of utterance and token numbers per dialogue across \datasetname and other dialogical datasets listed in Table~\ref{tab:stats_compare}.}
    \vspace{-3mm}
    \label{fig:ccdf_token}
\end{figure*}

Using these strategies on annotation results of question $\mathcal{Q}1$, we obtained two versions of \datasetname datasets. 
We denote the dataset generated using consensus agreement as \datasetnamecon and the one using majority agreement as \datasetnamemaj.

We employed a specific strategy on both \datasetnamecon and \datasetnamemaj to generate the final labels for manipulative techniques and targeted vulnerabilities.
If a technique or vulnerability is annotated by at least two annotators in one task, the technique or vulnerability will be added as the answer.
This resulted in some dialogues lacking technique and vulnerability labels.

\subsection{Dataset Statistics}\label{subsec:dataset_statistics}
In this section, we delve into the statistics of our datasets, \datasetnamecon and \datasetnamemaj, as depicted in Table~\ref{tab:final_label} and illustrated through Figures~\ref{fig:mentalmanip_stats1}, \ref{fig:heat_map}, and \ref{fig:embedding_space}. 
Our analysis utilizes a multi-class sentiment classification model, specifically the Distilbert-base-uncased-emotion model from Hugging Face, to determine the dominant emotion within each dialogue.

The analysis, presented in the left two panels of Figure~\ref{fig:mentalmanip_stats1}, indicates a strong alignment in the distribution of manipulation techniques and vulnerabilities between \datasetnamecon and \datasetnamemaj.
Additionally, the same figure's right panel reveals that both manipulative and non-manipulative dialogues within \datasetnamecon exhibit similar emotional distributions, with ``joy'' and ``anger'' being the two most common emotions. Figure~\ref{fig:heat_map} offers a heat map that elucidates the correlation between manipulation techniques and vulnerabilities, uncovering prevalent patterns like the association of accusations with shaming or belittling. Moreover, Figure~\ref{fig:embedding_space}'s left panel showcases a t-SNE visualization of Sentence Transformer embeddings for both manipulative and non-manipulative dialogues within \datasetnamecon, using the all-MiniLM-L12-v2 model from Hugging Face. This visualization underscores the difficulty of distinguishing between manipulative and non-manipulative dialogues due to their intertwined embeddings.

Furthermore, we compare \datasetname with other dialogical datasets listed in Table~\ref{tab:stats_compare}, noting that \datasetname encompasses a greater volume of conversational exchanges, suggesting a richer dialogue context. The Complementary Cumulative Distribution Function (CCDF) for utterance and token counts of \datasetname compared to other dialogical datasets is depicted in Figure~\ref{fig:ccdf_token}. The right panel of Figure~\ref{fig:embedding_space} visualizes the distribution of these datasets in the embedding space, illustrating significant overlap among them, except for the distinct clustering pattern of Fox News comments.

In summary, our analysis highlights the challenge of differentiating between manipulative and non-manipulative dialogues, indicating that reliance on emotion classification or conventional text embeddings alone is insufficient for this purpose. Moreover, our dataset's comparison with other datasets confirms its comprehensive distribution and diversity, aligning with the variety observed in related datasets.

%% file: tables/taxonomy_tree.tex

\forestset{
  folders down/.style={
    for tree={
      draw,
      rounded corners=false,
      grow'=0,
      folder,
      inner ysep=-0.1mm,
      font=\fontsize{7.5pt}{10pt}\selectfont
    },
    for current and siblings={anchor=north west, child anchor=north},
    for descendants={
      draw=none,
      inner xsep=0.5mm,
      inner ysep=-0.8mm,
    },
  },
  myFork/.style={
    for tree={
      rounded corners,
      draw,
      fit=tight,
      align=center,
      inner ysep=-0.0mm,
      parent anchor=south,
    },
  },
}

\begin{figure}[t]
\tiny
\centering
\begin{forest} 
    myFork,
    [Label Taxonomy of \datasetname, font=\fontsize{8pt}{10pt}\selectfont, minimum width=4.8cm
        [\begin{tabular}{@{}c@{}}Presence of\vspace{-1mm}\\Manipulation\end{tabular}, 
        folders down, 
        minimum width=2.2cm
            [Manipulative]
            [Non-manipulative]
        ]
        [\begin{tabular}{@{}c@{}}Manipulation\vspace{-1mm}\\Technique\end{tabular}, 
        folders down, 
        minimum width=2.2cm
            [Denial]
            [Evasion]
            [Feigning Innocence]
            [Rationalization]
            [Playing the Victim Role]
            [Playing the Servant Role]
            [Shaming or Belittlement]
            [Intimidation]
            [Brandishing Anger]
            [Accusation]
            [Persuasion or Seduction]
        ]
        [\begin{tabular}{@{}c@{}}Targeted\vspace{-1mm}\\Vulnerability\end{tabular}, 
        folders down,
        minimum width=2.2cm
            [Over-responsibility]
            [Over-intellectualization]
            [Naivete]
            [Low self-esteem]
            [Dependency]
        ]
    ]
\end{forest}
\caption{Multi-level taxonomy of \datasetname.}
\vspace{-4mm}
\label{tree:taxonomy_tree}
\end{figure}

%% file: tables/phrase_matching.tex
{
\renewcommand{\arraystretch}{0.8}
\begin{table}[t!]
    \centering
    \resizebox{\columnwidth}{!}{%
    \begin{tabular}{l|c|c|c|c}
    \toprule
    Key Phrase Length $l$ & $<=4$ & $<=6$ & $<=10$ & $>10$ \\ 
    \midrule
    Matching Percentage $P$ & $100\%$ & $90\%$ & $80\%$ & $70\%$ \\
    \bottomrule
    \end{tabular}
    }
    \vspace{-2mm}
    \caption{Length-adaptive matching criterion.}
    \vspace{-4mm}
    \label{tab:phrase_matching}
\end{table}
}

%% file: tables/final_label.tex
\begin{table}[t]
    \setlength{\tabcolsep}{2pt}
    \centering
    \small
    \resizebox{\columnwidth}{!}{%
    \begin{tabular}{ccccc}
    \toprule
    Dataset  & \#Dialogue & Manip:Non-manip & Tech\% & Vul\% \\
    \toprule
    \datasetnamecon  & $2,915$ & $2.24:1$ & $60.0\%$ & $20.8\%$ \\
    \midrule
    \datasetnamemaj & $4,000$ & $2.38:1$ & $53.9\%$ & $18.3\%$ \\
    \bottomrule
    \end{tabular}
    }
    \vspace{-1mm}
    \caption{\textcolor{\highlightcolor}{Statistics of \datasetnamecon and \datasetnamemaj}, detailing dialogue counts (\#Dialogue), the manipulative to non-manipulative dialogue ratio, and the percentages of dialogues labeled with techniques (Tech\%) and vulnerability (Vul\%). The exact numbers are provided in Table~\ref{tab:final_label_2} in Appendix~\ref{appendix:dataset_statistics}.}
    \vspace{-4mm}
    \label{tab:final_label}
\end{table}

%% file: 04_Experiment.tex
\section{Experiments} \label{sec:experiment}
\subsection{Experiment Setting}
We conducted experiments of three classification tasks on \datasetnamecon and \datasetnamemaj to assess performance of state-of-art models in detecting mental manipulation: Manipulation Detection (Section~\ref{sec:exp_detection}), Technique Classification (Section~\ref{sec:exp_classification}), and Vulnerability Classification (Section~\ref{sec:exp_classification}). 
We analyzed the performance of GPT-4 Turbo~\cite{bubeckSparksArtificialGeneral2023}, Llama-2-7B, Llama-2-13B\footnote{Both Llama-2-7B and Llama-2-13B are Chat versions.}~\cite{touvronLLaMAOpenEfficient2023}, and RoBERTa-base~\cite{liuRoBERTaRobustlyOptimized2019} across three experimental settings: zero-shot prompting, few-shot prompting, and fine-tuning. 
For zero-shot prompting, we presented a dialogue to LLMs to assess if it contained elements of mental manipulation.
In few-shot prompting, aside from instructions, we randomly provided one non-manipulative and two manipulative dialogues with true answers as examples.
In fine-tuning, Llama-2-13B and RoBERTa-base were fine-tuned on specific datasets, with Llama-2-13B undergoing instruction tuning and RoBERTa-base receiving traditional supervised fine-tuning.
Formats for zero-shot and few-shot prompting are detailed in Appendix~\ref{appendix:prompt}.
For Llama's training on different datasets, instructions were adapted to fit respective tasks. 
GPT-4 Turbo's implementation followed OpenAI's official cookbook\footnote{\url{https://github.com/openai/openai-cookbook/}}.
Talkdown dataset was ignored due to its lengthy dialogues which far surpass the input token limit of RoBERTa-base.

For experiment data, we randomly split \datasetnamecon and \datasetnamemaj into training, validation, and test sets with a ratio of 6:2:2. 
We ensured proportional representation of manipulative and non-manipulative dialogues, and consistent inclusion of each technique and vulnerability across all sets. 
All experiments were performed on three Quadro RTX 6000 GPUs. 
We set the temperatures of GPT-4 Turbo and LLaMA-2 to $0.1$ and $0.6$, respectively.
\textcolor{\highlightcolor}{At these levels, the models' responses are more consistent and less random.}

We seek to elucidate the following aspects:
\begin{itemize}[noitemsep,topsep=2pt,parsep=2pt,partopsep=1pt]
    \item The effectiveness of LLMs in identifying and categorizing mental manipulation based on their inherent knowledge.
    \item The performance of LLMs when prompted with examples.
    \item The performance of LLMs post fine-tuning on relevant datasets.
\end{itemize}

\subsection{Manipulation Detection} \label{sec:exp_detection}
This task is framed as a binary classification task.
In our interactions with ChatGPT and GPT-4, we found it tends to mistakenly classify non-manipulative dialogues as manipulative if they feature general toxicity, like profanity, without actual manipulative intent.
Thus, we were keen to investigate the over-reactivity of LLMs when identifying mental manipulation.
\vspace{1mm}

\noindent\textbf{Hypersensitivity of LLMs}: We examined GPT-4 Turbo, Llama-2-7B, and Llama-2-13B on the manipulation detection task using \textcolor{\highlightcolor}{$899$} non-manipulative dialogues in \datasetnamecon.
Prediction results are detailed in Table~\ref{tab:exp_5}.
GPT-4 Turbo incorrectly identified $312$ dialogues as manipulative. 
Both Llama-2-7B and Llama-2-13B exhibited poor accuracy, mis-classifying almost all non-manipulative dialogues, with Llama-2-13B showing slightly better performance. 
These results indicate Llama-2's limited capability in accurately discerning mental manipulation.

\input{tables/exp_5}
\input{tables/exp_1}
\input{tables/exp_4}
\input{tables/exp_2}

Then, we conducted manipulation detection on the entirety of \datasetnamecon and \datasetnamemaj. 
Note that the distribution of manipulative and non-manipulative dialogues in both datasets is imbalanced, with manipulative dialogues being more prevalent, as detailed in Table~\ref{tab:final_label}.
We evaluated the models based on binary Precision, binary Recall, Accuracy, micro $F_1$, and macro $F_1$.
Because of binary classification, the accuracy has the same score as Micro $F_1$.

Experiment results are presented in Table~\ref{tab:exp_1} and Table~\ref{tab:exp_4}.
It is observed that \datasetnamemaj poses a greater challenge for prediction, as we expected.
In zero-shot and few-shot prompting, Llama-2-13B classifies nearly all dialogues as manipulative, causing high recall rates.
Few-shot prompting improves Accuracy and $F_1$ scores for both GPT-4 Turbo and Llama-2-13B.
For GPT-4 Turbo, few-shot prompting increases its Recall, making it more likely to identify dialogues as manipulative.
For Llama-2-13B, few-shot prompting makes it less sensitive and produces fewer manipulative predictions.
Appendix~\ref{appendix:confusion_matrix} provides the confusion matrices for prediction results of GPT-4 Turbo and Llama-2-13B under zero-shot and few-shot prompting on \datasetnamecon.
For fine-tuning, Llama-2-13B on Dreaddit gives the best performance among all finetuning results on existing datasets.
Note that Dreaddit is about detecting Mental Stress.
However, fine-tuning Llama-2-13B on all existing datasets does not notably enhance performance beyond zero-shot or few-shot prompting outcomes.
RoBERTa-base overall exhibits inferior Accuracy compared to Llama-2-13B.
Specifically, fine-tuning it on Fox News dataset results in badly degraded performance.
This decline may stem from the broader semantic distribution of the Fox News dataset, as illustrated in Figure~\ref{fig:embedding_space}.

\subsection{Technique and Vulnerability Classification} \label{sec:exp_classification}
Here we examined these models on multi-label classification tasks to identify manipulative techniques and victim vulnerabilities.
We present the experiment results on \datasetnamecon in Table~\ref{tab:exp_2}, where we report micro Precision and micro Recall.
For few-shot prompting on both classification tasks, we provided $2$ randomly chosen examples.
We can observe that GPT-4 Turbo performs better than Llama-2-13B under zero-shot and few-shot prompting, and few-shot prompting increases their accuracy.
Fine-tuning Llama-2-13B on \datasetnamecon still gives better performance than fine-tuning RoBERTa-base.
\textcolor{\highlightcolor}{Precision and recall scores for each technique and vulnerability category, specifically under zero-shot prompting for GPT-4 Turbo and Llama-2-13B, are provided in Table~\ref{tab:exp_3}.}

\begin{figure*}[t!]
\centering
    \begin{minipage}{0.4\textwidth}
        \centering
        \includegraphics[width=\linewidth]{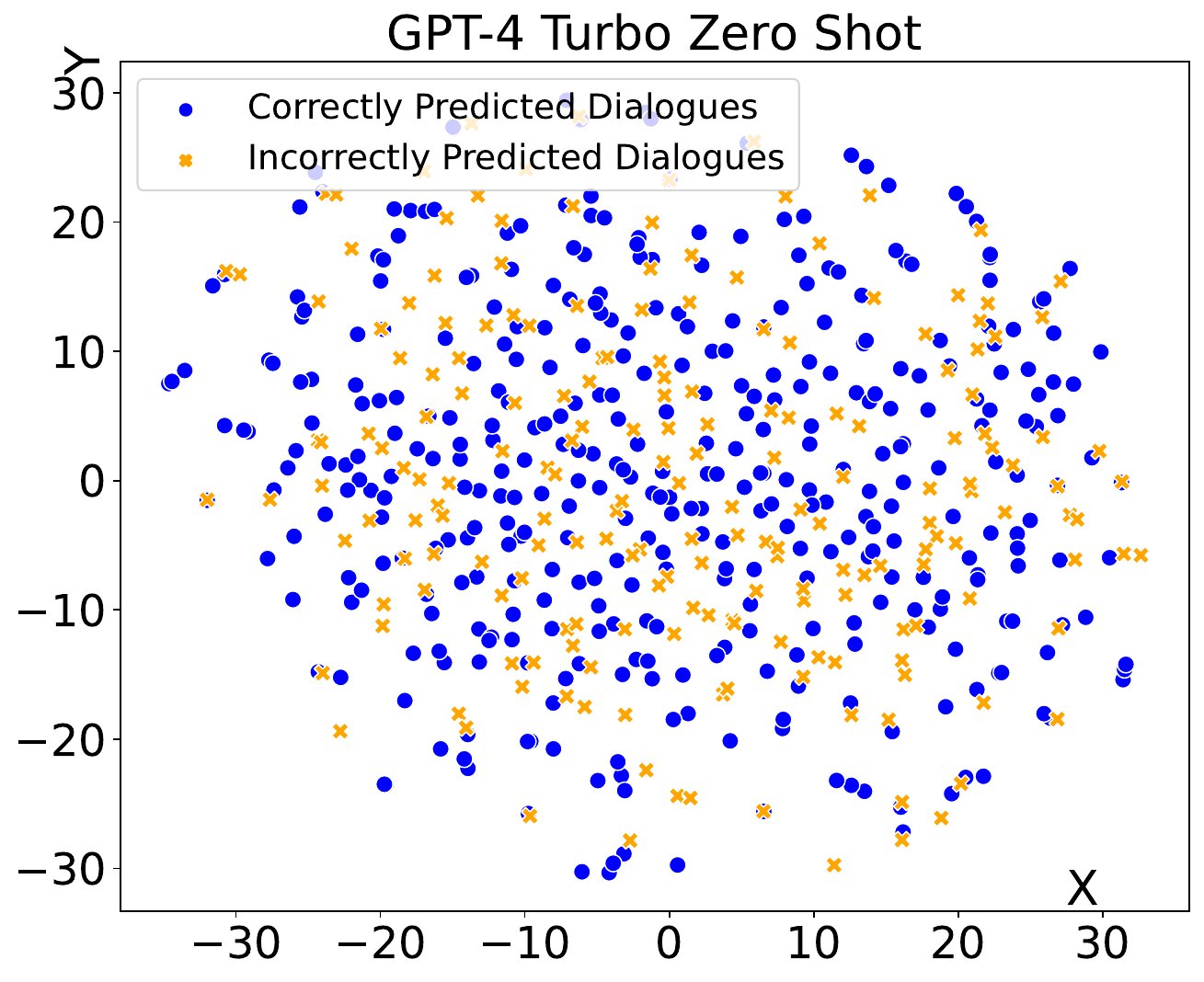}
    \end{minipage}
    \hspace{15pt}
    \begin{minipage}{0.4\textwidth}
        \centering
        \includegraphics[width=\linewidth]{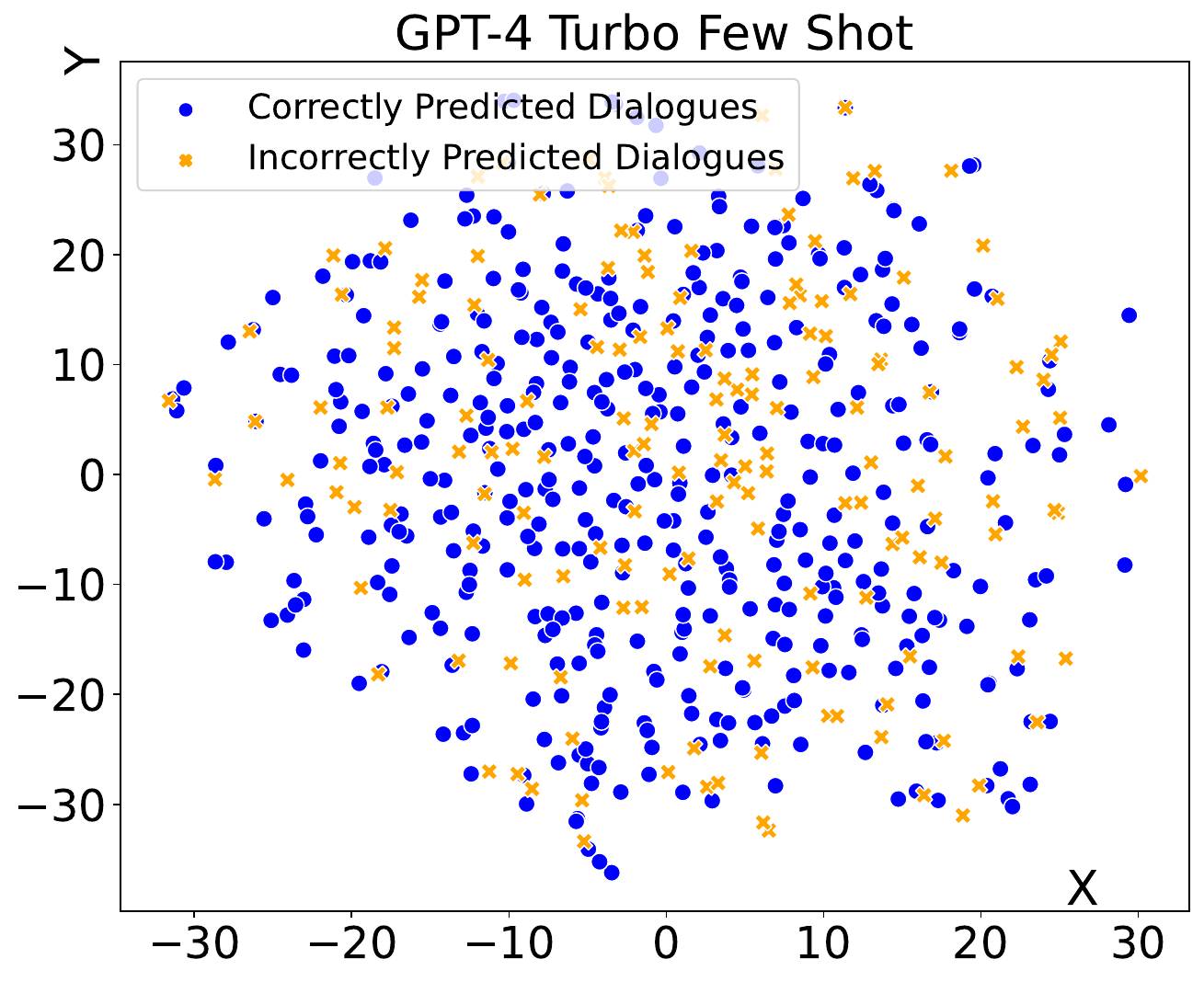}
    \end{minipage}
    \vspace{-1mm}
    \caption{\textcolor{\highlightcolor}{t-SNE visualization of Sentence Transformer embeddings of dialogues in \datasetnamecon's test set that are correctly or incorrectly predicted by GPT-4 Turbo under zero-shot setting and few-shot setting.}}
    \vspace{-1mm}
    \label{fig:negative_results}
\end{figure*}

\subsection{Analysis on Incorrect Predictions}
\textcolor{\highlightcolor}{
In this part, we investigate whether the dialogue instances where the models successfully predicted the correct label and those where they failed are semantically significantly different.
To do this, we extract dialogues from \datasetnamecon's test set that were correctly and incorrectly predicted by GPT-4 Turbo.
Under zero-shot setting, GPT-4 Turbo correctly predicted $383$ dialogues and incorrectly predicted $200$ ones.
Under few-shot setting, GPT-4 Turbo correctly predicted $422$ dialogues and incorrectly predicted $161$ ones.
Using t-SNE visualization of Sentence Transformer embeddings of both kinds of dialogues, we present their semantic distributions in Figure~\ref{fig:negative_results}.
The visualizations indicate that the two kinds of dialogues are semantically indistinguishable, highlighting the challenge of distinguishing manipulative language based solely on lexicon or semantic features.}

\input{tables/exp_3}

%% file: tables/exp_5.tex
{
\renewcommand{\arraystretch}{0.8}
\begin{table}[t!]
    \setlength{\tabcolsep}{2pt}
    \centering
    \small
    \resizebox{\columnwidth}{!}{%
    \begin{tabular}{lrrr}
     \toprule
      \textbf{Predictions} & \textbf{GPT-4 Turbo} & \textbf{Llama-2-7B} & \textbf{Llama-2-13B} \\
     \toprule
     Manipulative & $~~~312$ & $~~~895$ & $~~~879$ \\
     \midrule
     Non-manipulative & $~~~587$ & $~~~~~~~4$ & $~~~~~20$ \\
     \specialrule{0.3pt}{2pt}{0pt}
     \specialrule{0.3pt}{1pt}{3pt}
     \textbf{Accuracy} & $0.653$ & $0.004$ & $0.022$ \\
     \bottomrule
    \end{tabular}
    }
    \vspace{-2mm}
    \caption{\textcolor{\highlightcolor}{Out of $899$ non-manipulative dialogues in \datasetnamecon, the number of dialogues predicted as manipulative and non-manipulative.}}
    \vspace{-4mm}
    \label{tab:exp_5}
\end{table}
}

%% file: tables/exp_1.tex
{
\renewcommand{\arraystretch}{0.8}
\begin{table*}[t]
    \setlength{\tabcolsep}{2.5pt}
    \centering
    \small
    \resizebox{\textwidth}{!}{%
    \begin{tabular}{clccccccccccccccc}
    \toprule
    \multirow{2}{*}{Experiment Setting} & \multirow{2}{*}{Training Dataset} & \multicolumn{5}{c}{GPT-4 Turbo} & \multicolumn{5}{c}{Llama-2-13B} & \multicolumn{5}{c}{RoBERTa-base}
    \\ \cmidrule(lr){3-7}\cmidrule(lr){8-12}\cmidrule(lr){13-17}
    & & $P$ & $R$ & $ACC$ & $F_1^{mi}$ & $F_1^{ma}$ & $P$ & $R$ & $ACC$ & $F_1^{mi}$ & $F_1^{ma}$ & $P$ & $R$ & $ACC$ & $F_1^{mi}$ & $F_1^{ma}$
    \\ \toprule
    Zero-shot prompting &   & $.788$ & $.682$ & $.657$ & $.657$ & $.629$ & $.693$ & $.997$ & $.696$ & $.696$ & $.450$ & -- & -- & -- & -- & -- 
    \\ \midrule
    Few-shot prompting & \datasetnamecon & $.802$ & $.792$  & $.724$ & $.724$ & $.683$ & $.735$  & $.912$  & $.715$  & $.715$ & $.602$ & -- & -- & -- & -- & -- 
    \\ \midrule
    \multirow{11}{*}{Fine-tuning} 
    & Dreaddit & -- & -- & -- & -- & -- & $.721$ & $.982$ & $.727$ & $.727$ & $.559$ & $.864$ & $.208$ & $.435$ & $.435$ & $.422$
    \\ \cmidrule{2-17}
     & SDCNL & -- & -- & -- & -- & -- & $.698$  & $.995$ & $.702$ & $.702$ & $.471$ & $.684$ & $.822$ & $.619$ & $.619$ & $.488$ 
    \\ \cmidrule{2-17}
     & ToxiGen & -- & -- & -- & -- & -- & $.693$ & $.999$ & $.696$ & $.696$ & $.446$ & $.717$ & $.864$ & $.674$ & $.674$ & $.559$
    \\ \cmidrule{2-17}
     & DetexD & -- & -- & -- & -- & -- & $.696$ & $.992$ & $.698$ & $.698$ & $.465$ & $.803$ & $.215$ & $.427$ & $.427$ & $.416$
    \\ \cmidrule{2-17}
    & Fox News & -- & -- & -- & -- & -- & $.690$  & $.997$ & $.691$ & $.691$ & $.434$ & $.000$ & $.000$ & $.312$ & $.312$ & $.238$
    \\ \cmidrule{2-17}
    & ToxiChat & -- & -- & -- & -- & -- & $.689$ & $.999$ & $.691$ & $.691$ & $.429$ & $.791$ & $.333$ & $.483$ & $.483$ & $.483$
    \\ \cmidrule{2-17} 
    & MDRDC & -- & -- & -- & -- & -- & $.695$ & $.999$ & $.700$ & $.700$ & $.457$ & $.743$ & $.749$ & $.651$ & $.651$ & $.595$
    \\ \cmidrule{2-17}
    & \datasetnamecon & -- & -- & -- & -- & -- & $.828$ & $.835$ & $.768$ & $.768$ & $.731$ & $.786$ & $.904$ & $.766$ & $.766$ & $.700$
    \\ \bottomrule
    \end{tabular}
    }
    \vspace{-2mm}
    \caption{\textcolor{\highlightcolor}{Results of manipulation detection task on \textbf{\datasetnamecon}. $P$, $R$, $ACC$, $F_1^{mi}$, and $F_1^{ma}$ stands for binary precision, binary recall, accuracy, micro $F_1$, and macro $F_1$ respectively.}}
    \label{tab:exp_1}
\end{table*}
}

%% file: tables/exp_4.tex
{
\renewcommand{\arraystretch}{0.8}
\begin{table*}[t]
    \setlength{\tabcolsep}{2.5pt}
    \centering
    \small
    \resizebox{\textwidth}{!}{%
    \begin{tabular}{clccccccccccccccc}
    \toprule
    \multirow{2}{*}{Experiment Setting} & \multirow{2}{*}{Training Dataset} & \multicolumn{5}{c}{GPT-4 Turbo} & \multicolumn{5}{c}{Llama-2-13B} & \multicolumn{5}{c}{RoBERTa-base}
    \\ \cmidrule(lr){3-7}\cmidrule(lr){8-12}\cmidrule(lr){13-17}
    & & $P$ & $R$ & $ACC$ & $F_1^{mi}$ & $F_1^{ma}$ & $P$ & $R$ & $ACC$ & $F_1^{mi}$ & $F_1^{ma}$ & $P$ & $R$ & $ACC$ & $F_1^{mi}$ & $F_1^{ma}$
    \\ \toprule
    Zero-shot prompting &   & $.816$ & $.632$ & $.632$ & $.632$ & $.602$ & $.722$ & $.997$ & $.721$ & $.721$ & $.432$ & -- & -- & -- & -- & -- 
    \\ \midrule
    Few-shot prompting & \datasetnamemaj & $.812$ & $.710$ & $.672$ & $.672$ & $.627$ & $.732$ & $.979$ & $.726$ & $.726$ & $.486$ & -- & -- & -- & -- & -- 
    \\ \midrule
    \multirow{11}{*}{Fine-tuning} 
    & Dreaddit & -- & -- & -- & -- & -- & $.742$ & $.960$ & $.731$ & $.731$ & $.533$ & $.814$ & $.191$ & $.386$ & $.386$ & $.378$
    \\ \cmidrule{2-17}
     & SDCNL & -- & -- & -- & -- & -- & $.726$ & $.983$ & $.720$ & $.720$ & $.458$ & $.696$ & $.565$ & $.510$ & $.510$ & $.459$
    \\ \cmidrule{2-17}
     & ToxiGen & -- & -- & -- & -- & -- & $.723$ & $.997$ & $.723$ & $.723$ & $.436$ & $.731$ & $.734$ & $.615$ & $.615$ & $.521$
    \\ \cmidrule{2-17}
     & DetexD & -- & -- & -- & -- & -- & $.727$ & $.988$ & $.724$ & $.724$ & $.460$ & $.792$ & $.225$ & $.400$ & $.400$ & $.396$
    \\ \cmidrule{2-17}
    & Fox News & -- & -- & -- & -- & -- & $.722$ & $.997$ & $.721$ & $.721$ & $.432$ & $.000$ & $.000$ & $.280$ & $.280$ & $.218$
    \\ \cmidrule{2-17}
    & ToxiChat & -- & -- & -- & -- & -- & $.721$ & $.998$ & $.721$ & $.721$ & $.428$ & $.797$ & $.348$ & $.467$ & $.467$ & $.466$
    \\ \cmidrule{2-17} 
    & MDRDC & -- & -- & -- & -- & -- & $.724$ & $.998$ & $.725$ & $.725$ & $.441$ & $.779$ & $.682$ & $.632$ & $.632$ & $.581$
    \\ \cmidrule{2-17}
    & \datasetnamemaj & -- & -- & -- & -- & -- & $.809$ & $.851$ & $.748$ & $.748$ & $.673$ & $.791$ & $.875$ & $.743$ & $.743$ & $.651$
    \\ \bottomrule
    \end{tabular}
    }
    \vspace{-2mm}
    \caption{\textcolor{\highlightcolor}{Results of manipulation detection task on \textbf{\datasetnamemaj}. $P$, $R$, $ACC$, $F_1^{mi}$, and $F_1^{ma}$ stands for binary precision, binary recall, accuracy, micro $F_1$, and macro $F_1$ respectively.}}
    \label{tab:exp_4}
\end{table*}
}

%% file: tables/exp_2.tex
{
\renewcommand{\arraystretch}{0.8}
\begin{table*}[h!]
    \centering
    \small
    \begin{tabular}{cccccccccccc}
    \toprule
    \multirow{2}{*}{Experiment Setting} & \multirow{2}{*}{Model} & \multicolumn{5}{c}{Technique} & \multicolumn{5}{c}{Vulnerability}
    \\ \cmidrule(lr){3-7}\cmidrule(lr){8-12}
    & &  $P^{mi}$ & $R^{mi}$ & $ACC$ & $F_1^{mi}$ & $F_1^{ma}$ & $P^{mi}$ & $R^{mi}$ & $ACC$ & $F_1^{mi}$ & $F_1^{ma}$ 
    \\ \toprule
    \multirow{2}{*}{Zero-shot prompting} & GPT-4 Turbo & $.311$ & $.618$ & $.111$ & $.414$ & $.376$ & $.373$ & $.786$ & $.092$ & $.506$ & $.423$
    \\ \cmidrule(lr){2-12}
    & Llama-2-13B & $.174$ & $.448$ & $.025$ & $.250$ & $.233$ & $.164$ & $.366$ & $.000$ & $.227$ & $.222$
    \\ \midrule
    \multirow{2}{*}{Few-shot prompting} & GPT-4 Turbo & $.387$ & $.533$ & $.224$ & $.449$ & $.394$ & $.429$ & $.626$ & $.269$ & $.509$ & $.370$
    \\ \cmidrule(lr){2-12}
    & Llama-2-13B & $.324$ & $.283$ & $.205$ & $.302$ & $.193$ & $.157$ & $.183$ & $.042$ & $.169$ & $.162$
    \\ \midrule
    \multirow{2}{*}{Fine-tuning} & Llama-2-13B & $.349$ & $.821$ & $.029$ & $.490$ & $.384$ & $.265$ & $.756$ & $.008$ & $.393$ & $.280$
    \\ \cmidrule(lr){2-12}
    & RoBERTa-base & $.479$ & $.470$ & $.264$ & $.475$ & $.334$ & $.532$ & $.496$ & $.445$ & $.513$ & $.250$
    \\ \bottomrule
    \end{tabular}
    \vspace{-2mm}
    \caption{Results of technique and vulnerability multi-label classification on \textbf{\datasetnamecon}. $P^{mi}$, $R^{mi}$, $ACC$, $F_1^{mi}$ and $F_1^{ma}$ stands for micro precision, micro recall, accuracy, micro $F_1$ and macro $F_1$, respectively.}
    \vspace{-4mm}
    \label{tab:exp_2}
\end{table*}
}

%% file: tables/exp_3.tex
\begin{table}[t!]
    \setlength{\tabcolsep}{5pt}
    \centering
    \small
    \resizebox{\columnwidth}{!}{%
    \begin{tabular}{lcccc}
    \toprule
    \multirow{2}{*}{Technique/Vulnerability} & \multicolumn{2}{c}{GPT-4 Turbo} & \multicolumn{2}{c}{Llama-2-13B}
    \\ \cmidrule(lr){2-3}\cmidrule(lr){4-5}
    & $P$ & $R$ & $P$ & $R$
    \\ \toprule
    Denial &  $0.180$ & $0.857$ & $0.085$  & $0.900$
    \\ 
    Evasion & $0.208$ & $0.714$ & $0.060$ & $1.000$
    \\ 
    Feigning Innocence & $0.184$ & $0.823$ & $0.063$ & $0.563$
    \\ 
    Rationalization & $0.204$ & $0.789$ & $0.178$ & $0.568$
    \\ 
    Playing the Victim Role & $0.056$ & $0.875$ & $0.071$ & $0.625$
    \\ 
    Playing the Servant Role & $0.138$ & $1.000$ & $0.000$ & $0.000$
    \\ 
    Shaming or Belittlement & $0.473$ & $0.709$ & $0.304$ & $0.688$
    \\ 
    Intimidation & $0.476$ & $0.861$ & $0.500$ & $0.467$
    \\ 
    Brandishing Anger & $0.538$ & $0.259$ & $0.208$ & $0.200$
    \\ 
    Accusation & $0.450$ & $0.529$ & $0.353$ & $0.358$
    \\
    Persuasion or Seduction & $0.778$ & $0.395$ & $0.610$ & $0.217$
    \\ \midrule
    Over-responsibility & $0.180$ & $0.692$ & $0.109$ & $1.000$
    \\ 
    Over-intellectualization & $0.200$ & $0.222$ & $0.136$ & $0.667$
    \\ 
    Naivete & $0.234$ & $0.833$ & $0.187$ & $0.944$
    \\  
    Low self-esteem & $0.384$ & $0.909$ & $0.200$ & $0.182$
    \\ 
    Dependency & $0.635$ & $0.810$ & $0.750$ & $0.103$
    \\ \bottomrule
    \end{tabular}
    }
    \vspace{-2mm}
    \caption{Precision and Recall of each technique and vulnerability category under zero-shot setting.}
    \vspace{-4mm}
    \label{tab:exp_3}
\end{table}

%% file: 05_Conclusion.tex
\section{Conclusion and Future Studies}
This study introduces \datasetname, a pioneering dataset aimed at identifying and classifying mental manipulation \textcolor{\highlightcolor}{in a fine-grained level.}
We assessed GPT-4 Turbo, Llama-2-13B, and RoBERTa-base across three classification tasks under various settings.
Experiment results reveal that models' understanding of mental manipulation do not align well with human perspectives. 
LLMs tend to incorrectly identify general toxicity as manipulation, a challenge particularly pronounced in smaller LLMs such as Llama-2-7B and Llama-2-13B. 
\textcolor{\highlightcolor}{In future work, it would be worthwhile to expand the dataset sources beyond the Cornell Movie Dialog Corpus and incorporate real-case interpersonal interaction data.
We also recognize that the performance of LLMs on more complex prompting paradigms, such as chain-of-thought, can be investigated.}

\textcolor{\highlightcolor}{Detecting and finely classifying mental manipulation in conversations is a challenging task due to the implicit and context-dependent nature of the language often used.
Furthermore, the subjectivity of human judgment complicates the alignment of AI models' predictions with human choices.
We have made the \datasetname dataset publicly available for future studies and hope it will inspire and foster further research in various NLP tasks and applications.}


%% file: 06_Limitation.tex
\section{\textcolor{\highlightcolor}{Limitations}}
We recognize that \datasetname dataset has several limitations:

\vspace{1mm}
\noindent \textbf{Language and Format} The \datasetname is limited to English-language content and focuses exclusively on dialogues between two individuals. Real-world interactions, however, are frequently more multifaceted. Consequently, this restriction may limit the dataset's applicability to more complex scenarios.

\vspace{1mm}
\noindent \textbf{Data Source} The \datasetname is derived from online movie scripts, which means the speech style presented may not accurately reflect natural, real-life communication.

\vspace{1mm}
\noindent \textbf{Data Annotation} The process of annotation is inherently subjective, which can introduce uncertainties in the precision of labeling. 
Additionally, the selection of annotators could lead to significant biases. 
For example, despite an imbalanced gender demographic among our annotators, we did not find notable differences in inter-annotator agreement across genders. 
We recognize that despite our best efforts, assembling an annotator pool that perfectly mirrors the general population remains a challenging endeavor.


%% file: 07_Ethics.tex
\section{Ethics and Broad Impact}
Before annotating, we noted that many dialogues, especially from R-rated movies, contained profanity that might upset annotators. 
To protect their well-being, we rephrased these instances into milder language while keeping the original context.
When recruiting annotators, we emphasized ensuring a diverse team in terms of race and gender.
Throughout the annotation phase, we actively encouraged annotator feedback, as summarized in Appendix~\ref{appendix:feedback}.

The \datasetname dataset contains a range of uncensored sensitive materials, including hate speech, violence, threats, mental health issues, sexual content, profanity, and more.
While our dataset is primarily designed for detection and classification tasks, we recognize the potential for misuse, particularly in the training of malicious generative AI systems. 
\textcolor{\highlightcolor}{For example, there is a risk that the data could be used to create automated chatbot systems that employ manipulative language for unethical purposes like scams.}
It is crucial to address these risks to ensure responsible use.

%% file: 08_Acknowledgment.tex
\section*{Acknowledgments}
We thank the Label Studio Platform for supporting student researchers by providing complimentary access to the Label Studio Enterprise Cloud Platform through their Academic Program.
We also thank the annotators for their time and effort.

%% file: appendix.tex
\section{Description of Taxonomy}
\label{appendix:definition}

Definitions of the $11$ manipulation techniques:
\begin{enumerate}[noitemsep]
    \item Denial~(DEN): The manipulator either denies wrongdoing or pretends to be confused about others' concerns.
    \item Evasion~(EVA): The manipulator refuses to pay attention to something or gives irrelevant or vague responses.
    \item Feigning innocence~(FEI): The manipulator implies that any harm caused was accidental.
    \item Rationalization~(RAT): The manipulator rationalizes their inappropriate behavior with excuses.
    \item Playing the Victim Role~(VIC): The manipulator portrays themselves as a victim to gain sympathy, attention, or divert focus from their misconduct.
    \item Playing the Servant Role~(SER): The manipulator disguises their self-serving motives as a contribution to a more noble cause.
    \item Shaming or Belittlement~(S\_B): The manipulator uses sarcasm, criticism, and put-downs to make others feel inferior, unworthy, or embarrassed.
    \item Intimidation~(INT): The manipulator places others on the defensive by using veiled threats.
    \item Brandishing Anger~(B\_A): The manipulator uses anger to brandish emotional intensity to shock the victim into submission.
    \item Accusation~(ACC): The manipulator suggests that the victim is at fault, selfish, uncaring, or leading an excessively easy life.
    \item Persuasion or Seduction~(P\_S): The manipulator employs charm, emotional appeal, or logical reasoning to make others lower their defenses.
\end{enumerate}

Definitions of the $5$ vulnerabilities targeted:
\begin{enumerate}[noitemsep]
    \item Naivety~(NAT): The victim is easily trusting and struggles to accept that others might be malevolent.
    \item Dependency~(DEP): The victim has interest-based or emotional dependencies on the manipulator.
    \item Over-responsibility~(O\_R): The victim is overly self-critical and sets high standards for themselves, often assuming undue blame and responsibility for the manipulator's actions.
    \item Over-intellectualization~(O\_I): The victim rationalizes the manipulator's hurtful behavior by believing there is always a justified reason behind it.
    \item Low self-esteem~(L\_S): The victim is self-doubting and unconfident in pursuing their own wants and needs.
\end{enumerate}

\begin{figure*}[t!]
\centering
    \begin{minipage}[b]{0.50\textwidth}
        \centering
        \includegraphics[width=\linewidth]{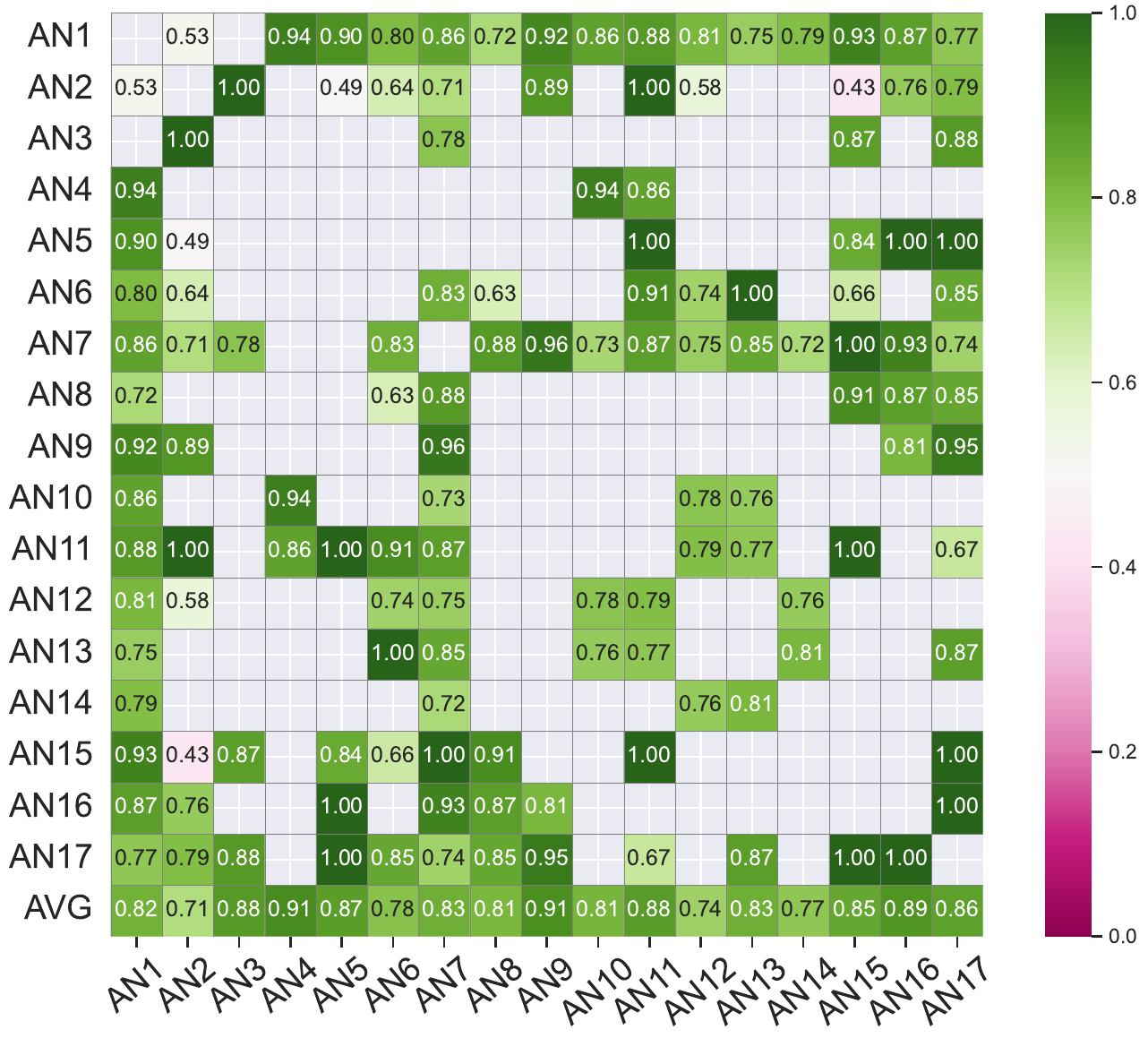}
    \end{minipage}
    \hspace{6pt}
    \begin{minipage}[b]{0.41\textwidth}
        \centering
        \includegraphics[width=\linewidth]{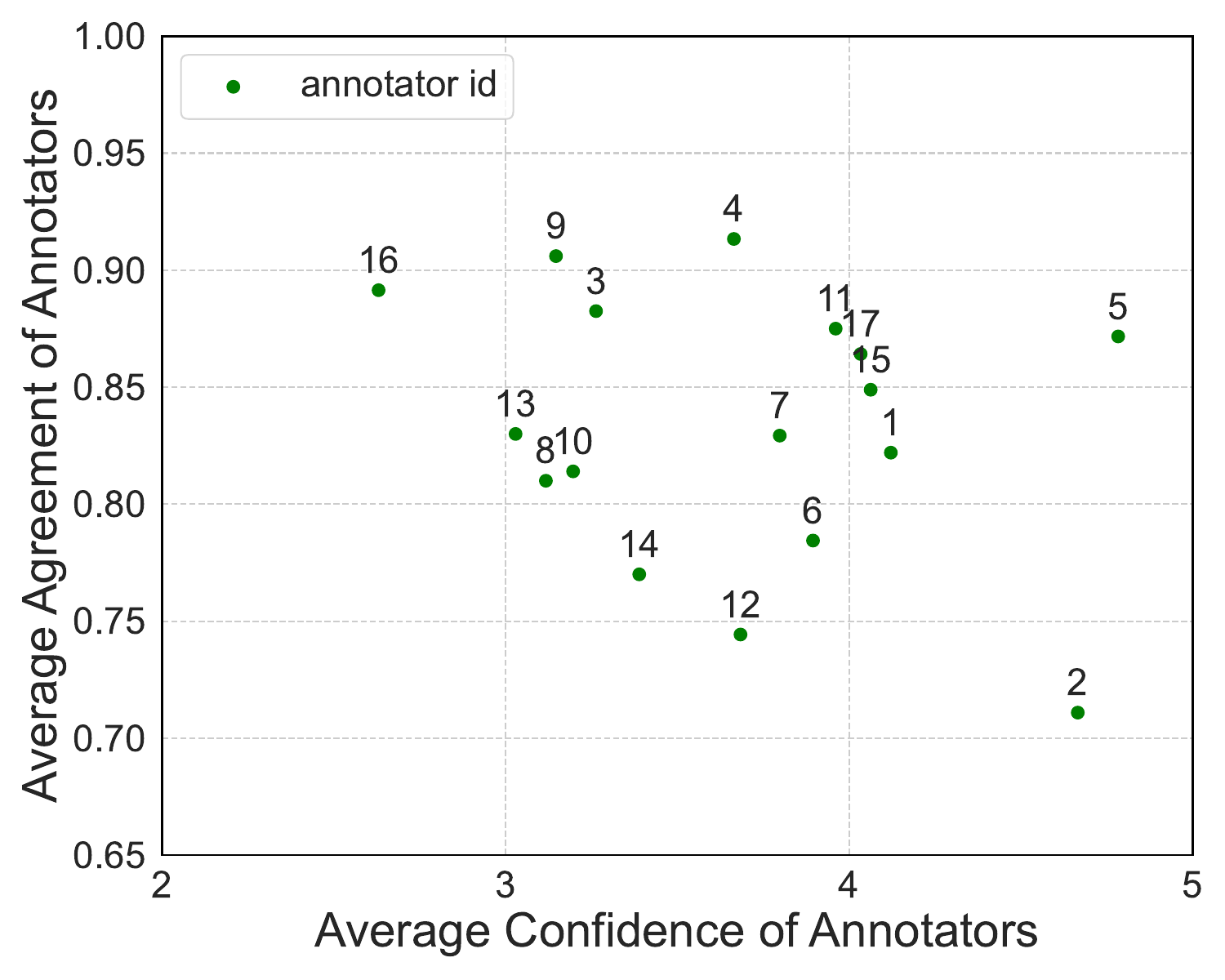}
    \end{minipage}
    \caption{\textcolor{\highlightcolor}{(left) Inter-annotator agreement of any two annotators based on their answers of whether a dialogue is manipulative. The last row is the average agreement score of each annotator. (right) Scatter plot of annotators' average confidence and inter-annotator agreement scores.}}
    \label{fig:agreement}
\end{figure*}

\begin{figure*}[h!]
\centering
    \begin{minipage}[b]{0.4\textwidth}
        \centering
        \includegraphics[width=\linewidth]{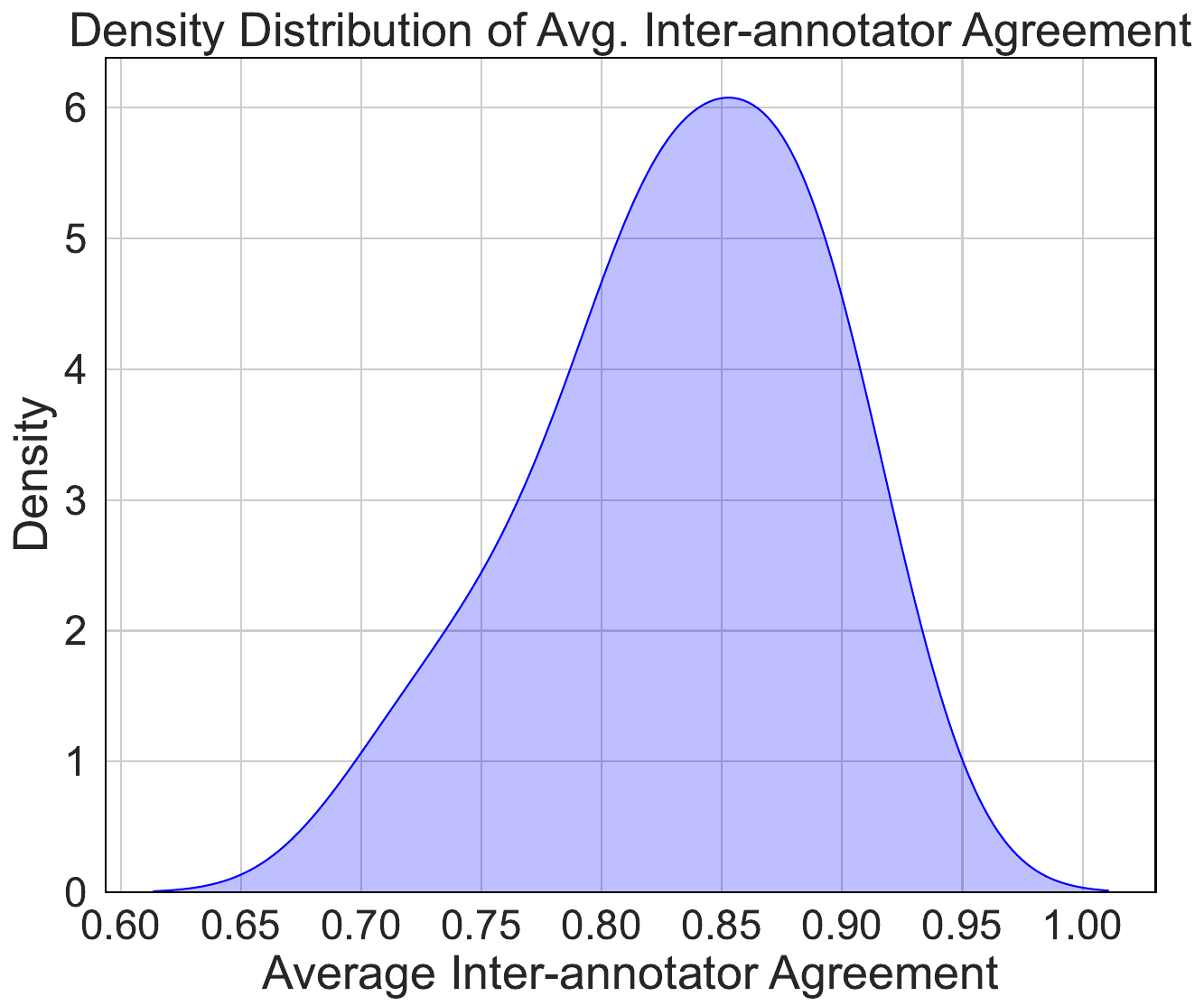}
    \end{minipage}
    \hspace{5pt}
    \begin{minipage}[b]{0.4\textwidth}
        \centering
        \includegraphics[width=\linewidth]{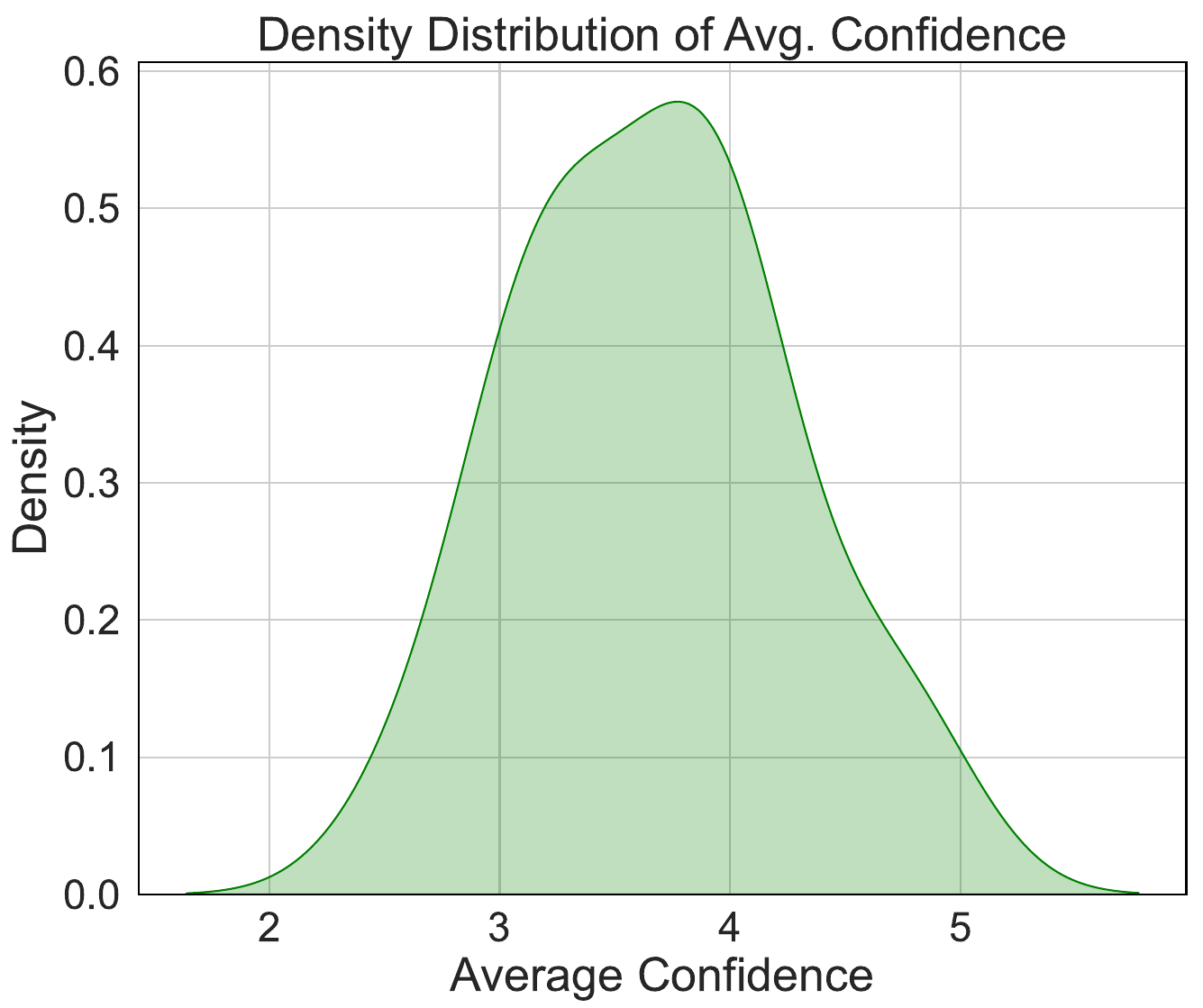}
    \end{minipage}
    \caption{\textcolor{\highlightcolor}{Density distribution of inter-annotator agreement and confidence of annotators.}}
    \label{fig:density_distribution}
\end{figure*}

\section{Key Phrase-based Matching} \label{appendix:key_phrase}
\subsection{Key Phrases Example}
\begin{itemize}[noitemsep]
    \item ``you make me do this''
    \item ``how could you do this to me''
    \item ``know your place''
    \item ``you should not feel that way''
    \item ``what more do you want''
    \item ``i do not remember''
    \item ``i do not like drama''
    \item ``watch your step''
    \item ``you always do this''
    \item ``you are too sensitive''
    \item ``it was not intentional''
    \item ``you do not love me''
    \item ``you would do it if you love me''
    \item ``it is all in the past''
\end{itemize}
The complete set of key phrases can be found in our GitHub repository.

\subsection{Online Resources}
We collected as many different manipulative phrases as we could from a number of websites suggested by Google by searching \textit{“phrases manipulative people use”} and likewise. 
Some of the websites browsed are below:
\begin{itemize}[noitemsep]
    \item https://geediting.com/10-phrases-manipulative-people-use-to-control-their-partner-in-a-relationship
    \item https://thevessel.io/phrases-manipulative-people-use-to-play-the-victim
    \item https://hackspirit.com/common-phrases-people-use-to-manipulate-you
    \item https://geediting.com/phrases-manipulative-people-use-to-blame-others-for-their-actions
\end{itemize}
We found that there was significant overlap across phrases as we collected phrases.

\section{Prompting Formats for GPT-4 and Llama-2}
\textcolor{\highlightcolor}{
When designing the prompts, we aimed to craft phrases that were both straightforward and broad, reflecting inquiries that real users might pose.}

\label{appendix:prompt}
\subsection{Manipulation Detection}
Zero-shot prompting format:
\begin{verbatim}
'''
I will provide you with a dialogue. 
Please determine if it contains elements 
of mental manipulation. Just answer with 
'Yes' or 'No', and don't add anything else.

<insert dialogue>
'''
\end{verbatim}

\noindent Few-shot prompting format:
\begin{verbatim}
'''
I will provide you with a dialogue. 
Please determine if it contains elements 
of mental manipulation. Just answer with 
'Yes' or 'No', and don't add anything else. 
Here are 3 examples:

Example 1:
<insert example_dialogue1>
<insert example_answer1>

Example 2:
<insert example_dialogue2>
<insert example_answer2>

Example 3:
<insert example_dialogue3>
<insert example_answer3>

<insert dialogue>
'''
\end{verbatim}

\subsection{Technique and Vulnerability Classification}
Zero-shot prompting format:
\begin{verbatim}
'''
Here are the definitions of 11 different 
mental manipulation techniques:
<insert definitions of 11 techniques>

Now, I will provide you with a dialogue 
that contains elements of mental manipulation.
Please determine which manipulative 
techniques are used by the manipulator.
Respond only with the names of the techniques,
and do not add anything else.

<insert dialogue>
'''
\end{verbatim}

\noindent Few-shot prompting format:
\begin{verbatim}
'''
Here are the definitions of 11 different 
mental manipulation techniques:
<insert definitions of 11 techniques>

Now, I will provide you with a dialogue 
that contains elements of mental manipulation.
Please determine which manipulative 
techniques are used by the manipulator.
Respond only with the names of the 
techniques, and do not add anything 
else. Here are 2 examples:

Example 1:
<insert example_dialogue1>
<insert example_answer1>

Example 2:
<insert example_dialogue2>
<insert example_answer2>

<insert dialogue>
'''
\end{verbatim}

For vulnerability classification, the prompting formats are similar.

\section{Analysis on Annotation Quality} \label{appendix:annotation_quality}
Figure~\ref{fig:agreement} is a heat map showing the inter-annotator agreement scores between any two annotators based on their answers for question $\mathcal{Q}$1.
For annotator $ANi$ and annotator $ANj$, their agreement score is calculated as:
\[ score_{ij} = \frac{\|\text{Annotations with same results}\|}{\| \text{Common tasks of } ANi \text{ and } ANj\| } \]
The last row is the average agreement score of each annotator.
We can see that all annotators have a moderate to strong average agreement ($\geq 0.7$) with other annotators assigned with the same tasks.

Figure~\ref{fig:agreement} is the scatter plot of annotators' average confidence and inter-annotator agreement scores. 
$16$ out of $17$ annotators reported an average confidence score above 3. 
We calculated the Spearman's rank correlation between annotators' inter-annotator agreement and confidence levels.
The statistic value is $-0.21$ and P-value is $0.41$, which reveals a very slight negative correlation between inter-annotator agreement and confidence levels, which is surprising for us.
This observation may be attributed to several factors.
Firstly, many annotators tend to assign lower or medium confidence scores (2 or 3) when they are not entirely certain of their decisions, regardless of their comprehension of the dialogue and the available options. 
Conversely, some annotators habitually assign high confidence scores (4 or 5) to most of their decisions, reflecting individual differences in confidence assessment.
Secondly, the assessment of mental manipulation is inherently subjective and lacks a uniform standard. 
Variations in what constitutes manipulation among annotators—with some setting higher thresholds than others—further diminish the reliability of inter-annotator agreement as a measure of annotation quality.

We also analyzed the inter-annotator agreement and average confidence by gender, as detailed in Table~\ref{tab:gender_agreement}. 
On average, male annotators exhibited higher confidence scores compared to their female counterparts. Furthermore, the inter-annotator agreement was higher among male annotators than among females. 
These differences could be significantly affected by the number of annotators of each gender and the volume and difficulty of tasks to which they were jointly assigned.

Figure~\ref{fig:density_distribution} illustrates the density distributions of annotators' average agreement and confidence scores, both of which exhibit a normalized distribution.

\input{tables/gender_agreement}

\section{\textcolor{\highlightcolor}{More Statistics of \datasetnamemaj}} \label{appendix:statistics_maj}
\begin{figure}[t!]
    \centering
    \includegraphics[width=0.85\linewidth]{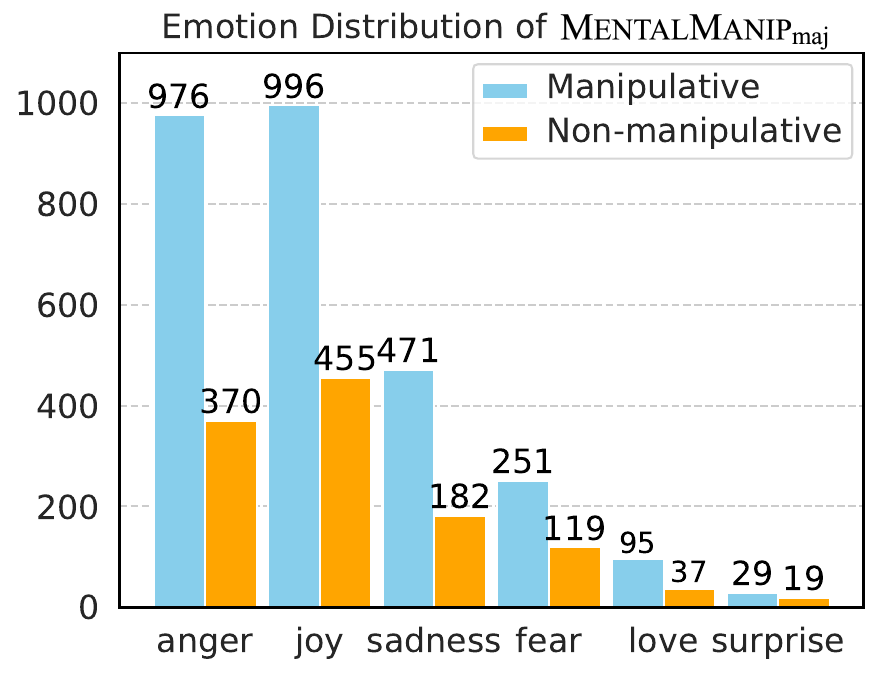}
    \caption{Emotion distribution of dialogues in dataset \datasetnamemaj.}
    \label{fig:senti_maj}
\end{figure}

\begin{figure*}[h!]
  \begin{minipage}{0.32\textwidth}
    \centering
    \includegraphics[width=\linewidth]{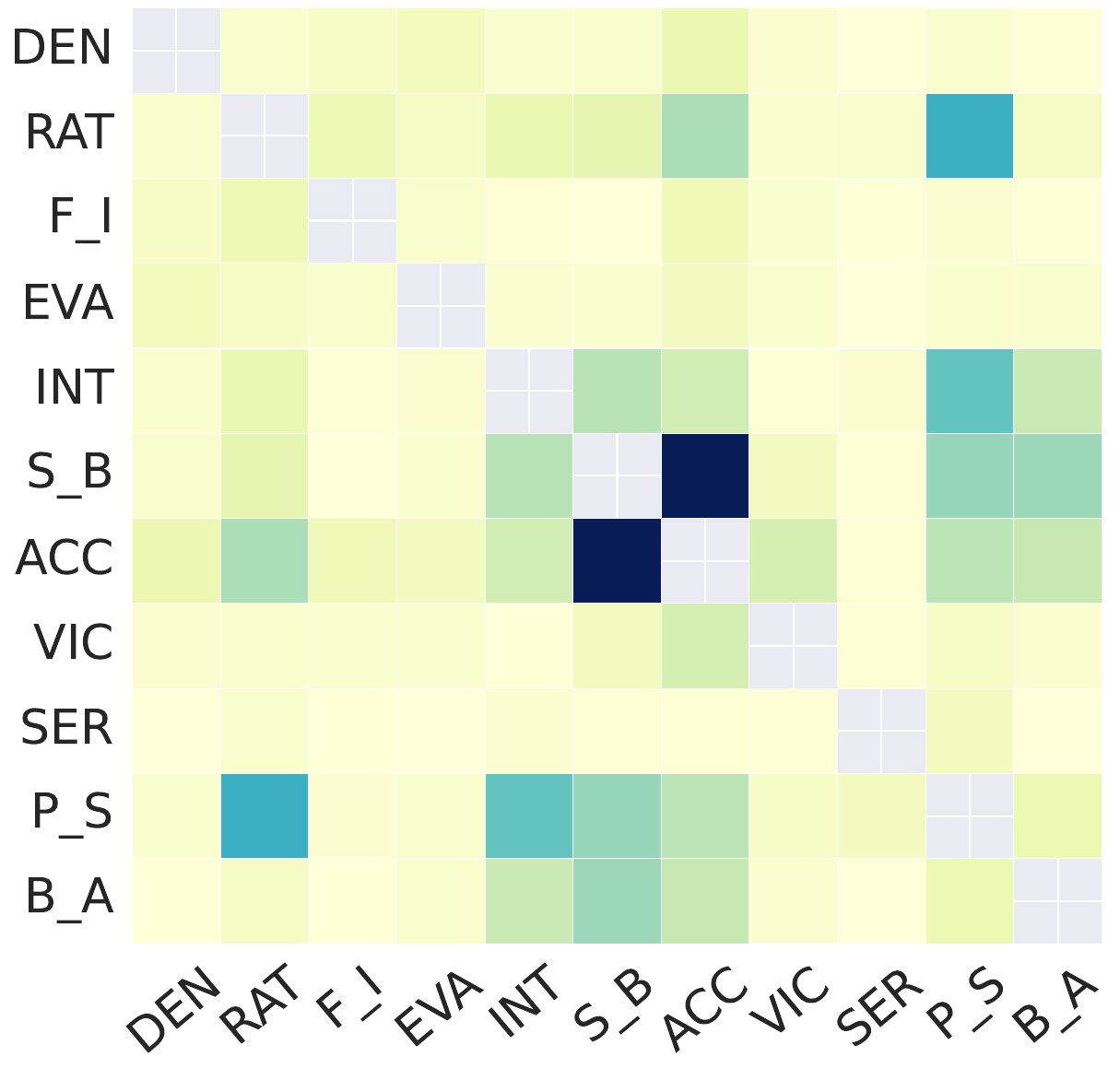}
  \end{minipage}
  \hspace{2pt}
  \begin{minipage}{0.32\textwidth}
    \centering
    \includegraphics[width=\linewidth]{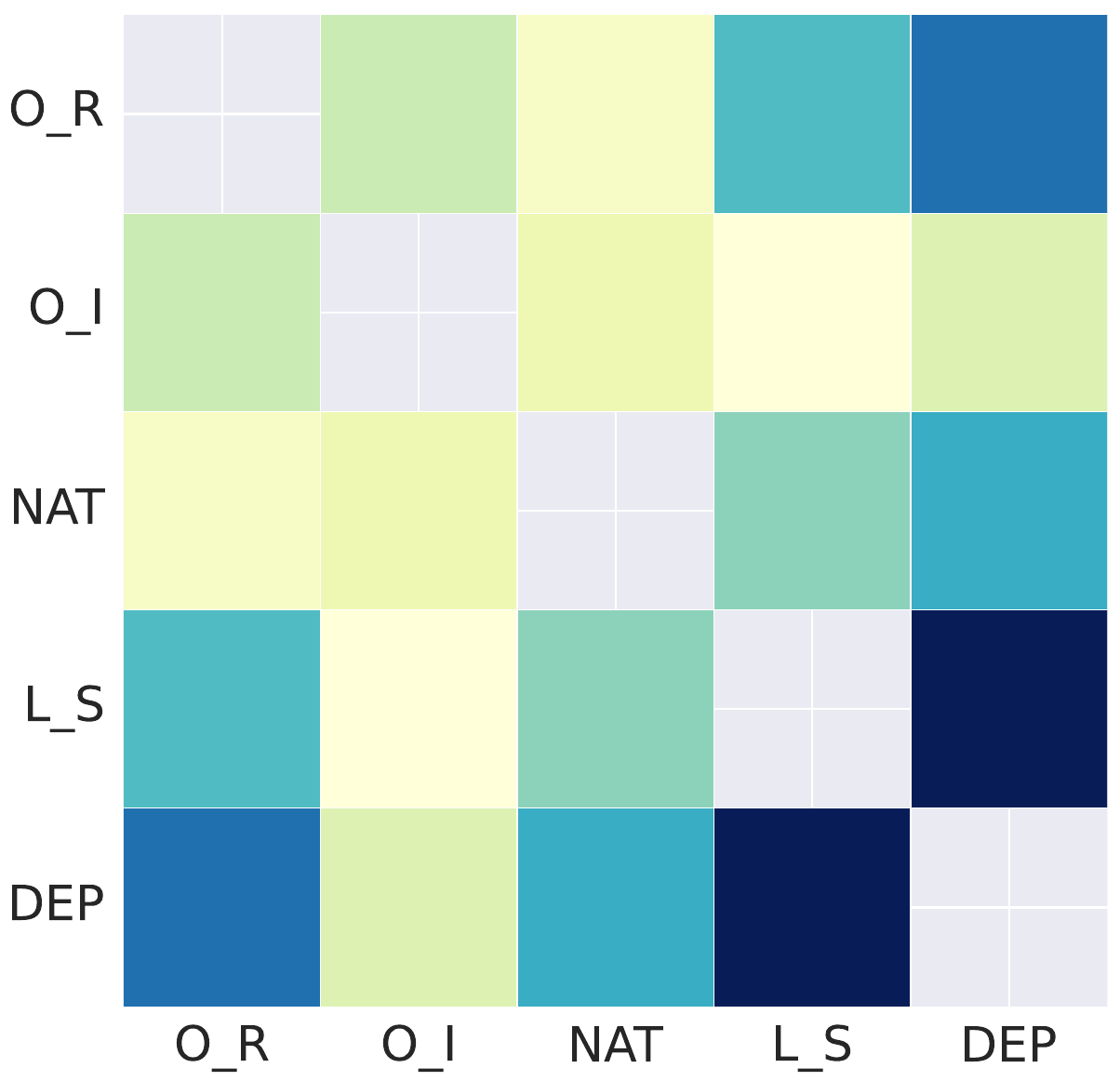}
  \end{minipage}
  \hspace{2pt}
  \begin{minipage}{0.32\textwidth}
    \centering
    \includegraphics[width=\linewidth]{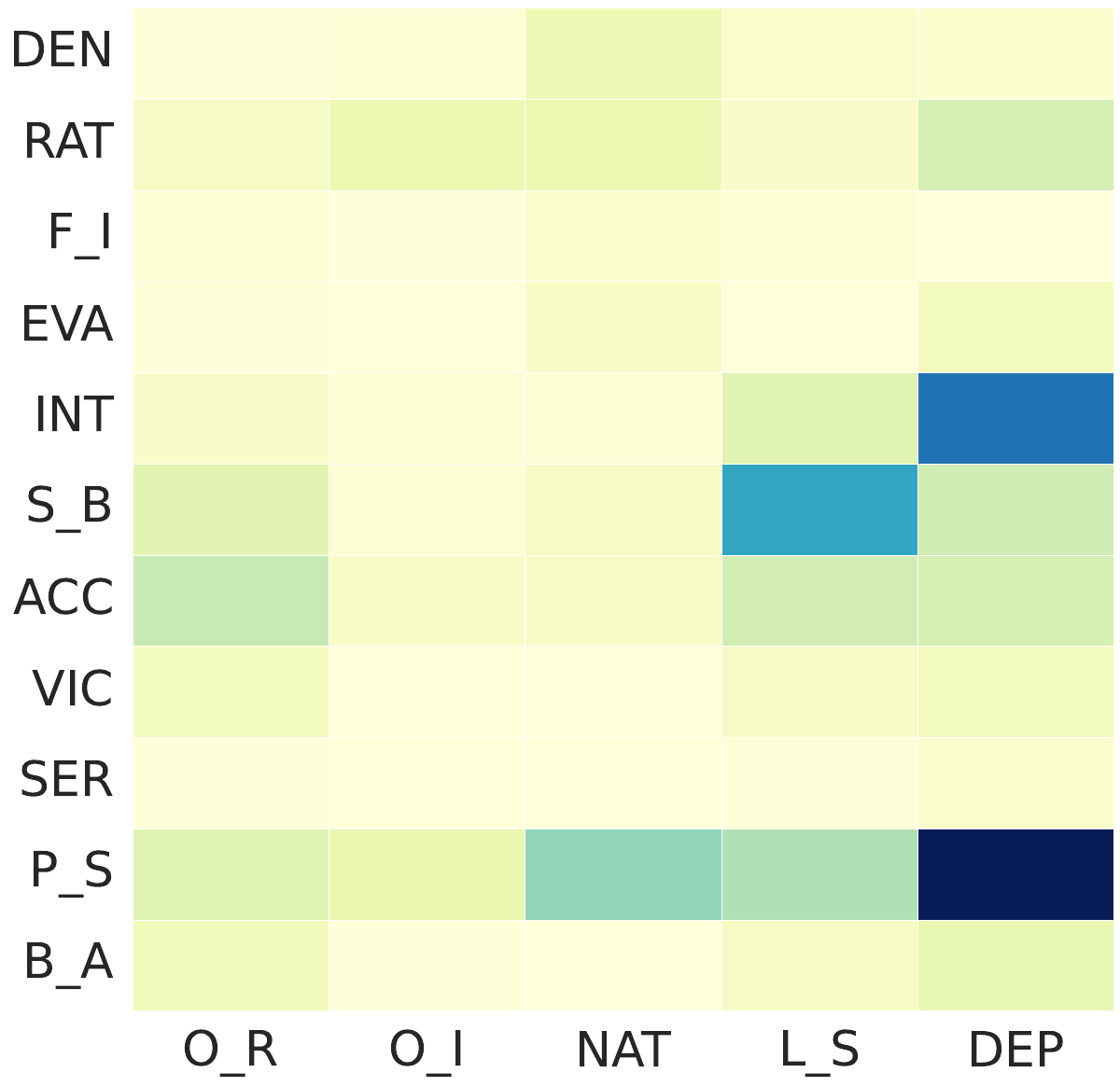}
  \end{minipage}
  \vspace{-1mm}
  \caption{Co-occurrence heat maps among techniques (left), vulnerabilities (center), and techniques and vulnerabilities (right) in \datasetnamemaj dataset. Darker cell indicates a higher co-occurrence.}
  \label{fig:heat_map_maj}
\end{figure*}

The emotion distribution of dialogues in \datasetnamemaj dataset is in Figure~\ref{fig:senti_maj}. 
The co-occurrence details are in Figure~\ref{fig:heat_map_maj}.

\section{\textcolor{\highlightcolor}{Confusion Matrices}} \label{appendix:confusion_matrix}
Please see Table~\ref{tab:conf_matrix_1}, \ref{tab:conf_matrix_2}, \ref{tab:conf_matrix_3}, and \ref{tab:conf_matrix_4}.

\begin{table}[h!]
\centering
\resizebox{\columnwidth}{!}{%
\begin{tabular}{|c|c|c|}\hline
\backslashbox{True Label}{Prediction} & Manipulative & Non-manipulative \\ 
\hline\hline
Manipulative & $272$  & $127$ \\
\hline
Non-manipulative & $73$  & $111$ \\ 
\hline
\end{tabular}
}
\vspace{-1mm}
\caption{Confusion matrix of \textbf{zero-shot} prompting result of GPT-4 Turbo on \textbf{\datasetnamecon}.}
\label{tab:conf_matrix_1}
\end{table}

\begin{table}[h!]
\centering
\resizebox{\columnwidth}{!}{%
\begin{tabular}{|c|c|c|}\hline
\backslashbox{True Label}{Prediction} & Manipulative & Non-manipulative \\ 
\hline\hline
Manipulative & $398$  & $1$ \\
\hline
Non-manipulative & $176$  & $8$ \\ 
\hline
\end{tabular}
}
\vspace{-1mm}
\caption{Confusion matrix of \textbf{zero-shot} prompting result of Llama-2-13B on \textbf{\datasetnamecon}.}
\label{tab:conf_matrix_2}
\end{table}

\begin{table}[h!]
\centering
\resizebox{\columnwidth}{!}{%
\begin{tabular}{|c|c|c|}\hline
\backslashbox{True Label}{Prediction} & Manipulative & Non-manipulative \\ 
\hline\hline
Manipulative & $316$  & $83$ \\
\hline
Non-manipulative & $78$  & $106$ \\ 
\hline
\end{tabular}
}
\vspace{-1mm}
\caption{Confusion matrix of \textbf{few-shot} prompting result of GPT-4 Turbo on \textbf{\datasetnamecon}.}
\label{tab:conf_matrix_3}
\end{table}

\begin{table}[h!]
\centering
\resizebox{\columnwidth}{!}{%
\begin{tabular}{|c|c|c|}\hline
\backslashbox{True Label}{Prediction} & Manipulative & Non-manipulative \\ 
\hline\hline
Manipulative & $382$  & $14$ \\
\hline
Non-manipulative & $126$  & $18$ \\ 
\hline
\end{tabular}
}
\vspace{-1mm}
\caption{Confusion matrix of \textbf{few-shot} prompting result of Llama-2-13B on \textbf{\datasetnamecon}.}
\label{tab:conf_matrix_4}
\end{table}

\input{tables/annotation_sample}


\input{annotation_observation}

\section{Annotation Example}
\label{appendix:annotation_example}
Table~\ref{tab:annotation_example} presents an annotation example.

\section{Annotation Platform and Instruction}
\label{appendix:screenshot}
\begin{figure*}[p]
  \centering
  \includegraphics[width=1.0\textwidth]{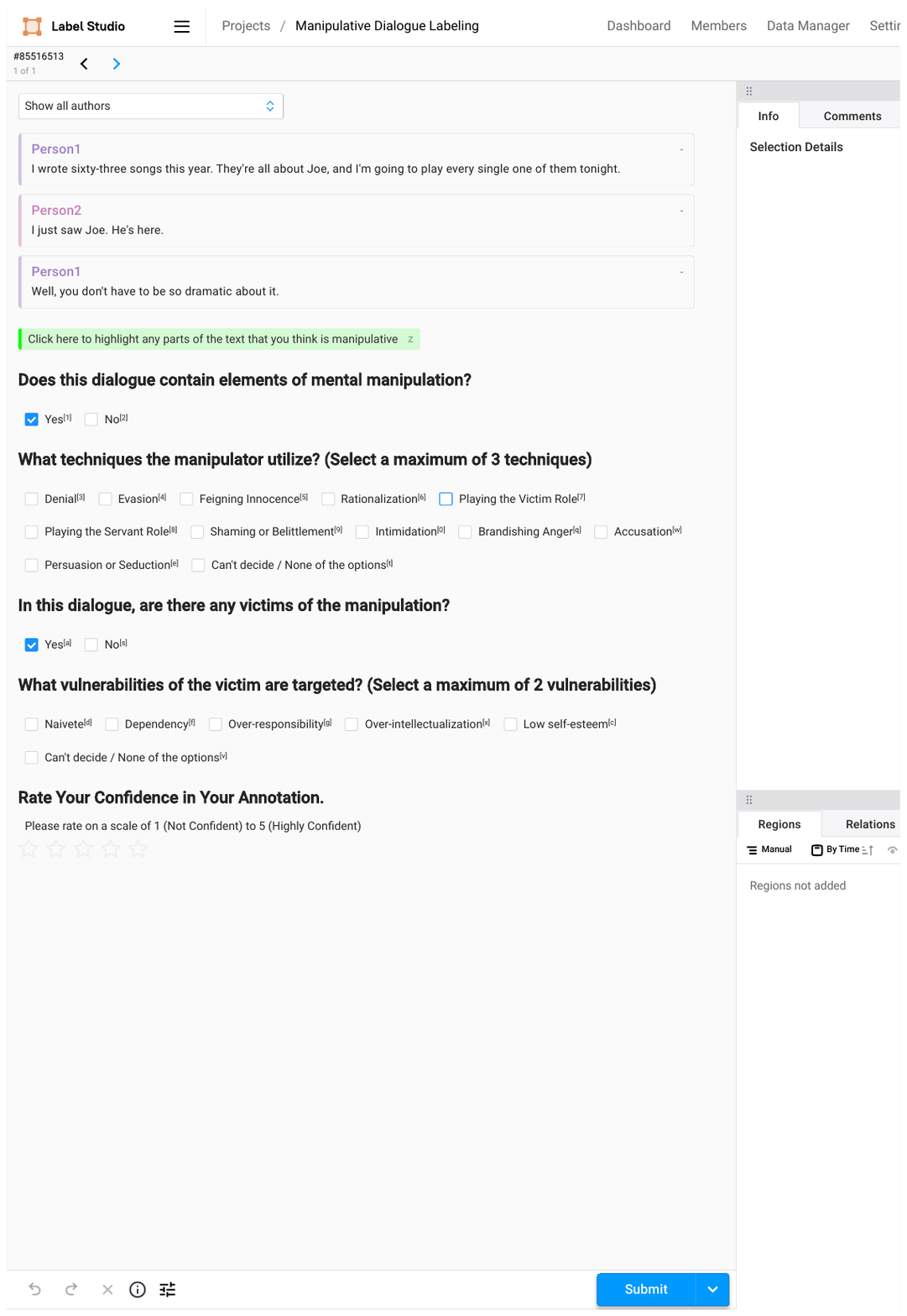}
  \caption{Screenshot of annotation platform interface.}
  \label{fig:annotation_interface}
\end{figure*}

\begin{figure*}[p]
  \centering
  \includegraphics[width=1.0\textwidth]{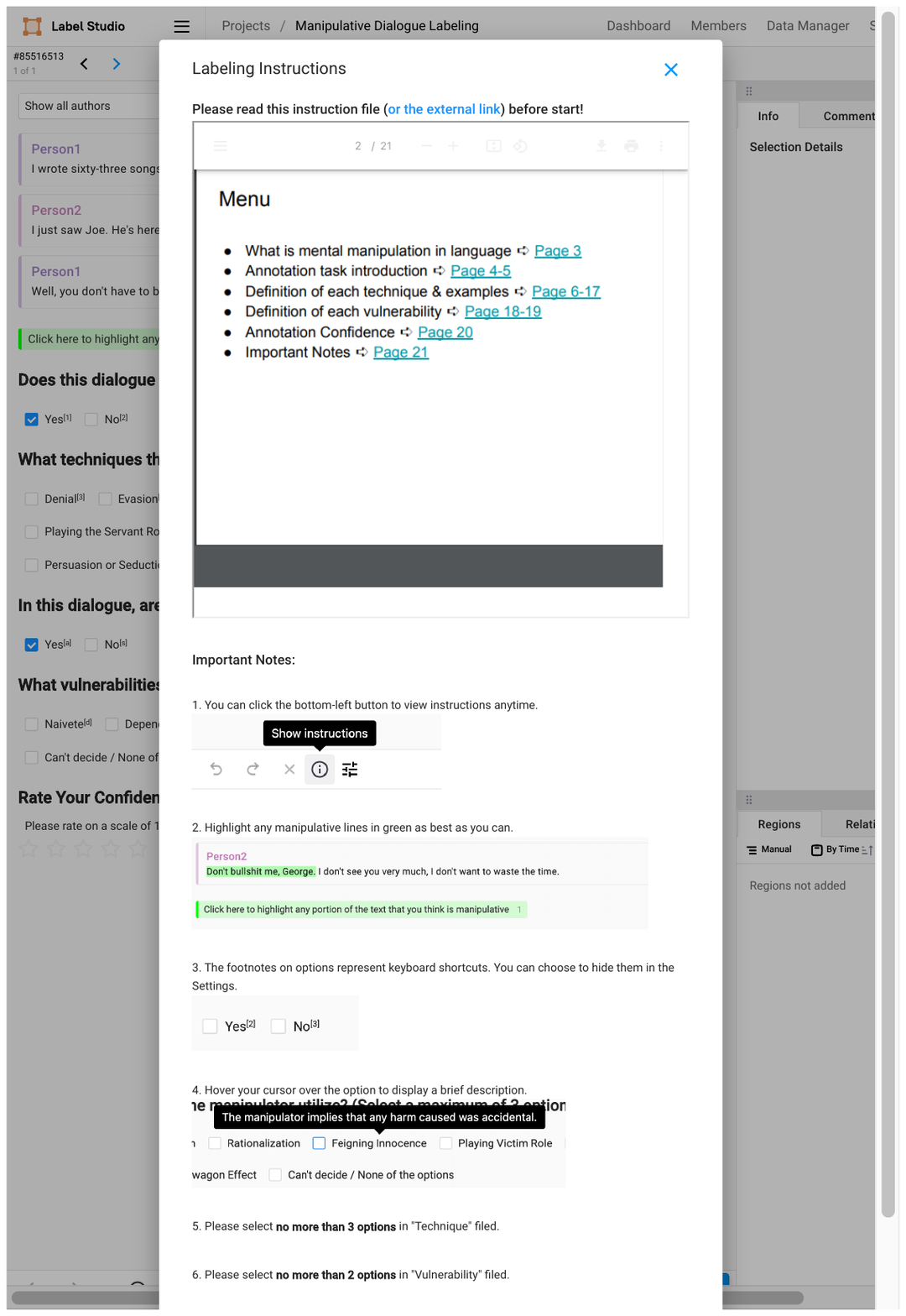}
  \caption{Screenshot of instruction window.}
  \label{fig:instruction_window}
\end{figure*}

Figure~\ref{fig:annotation_interface} is the screenshot of annotation platform interface, and Figure~\ref{fig:instruction_window} is the screenshot of instruction window.

\section{\textcolor{\highlightcolor}{More Dataset Statistics}} \label{appendix:dataset_statistics}
\begin{table}[h!]
    \setlength{\tabcolsep}{4pt}
    \centering
    \small
    \resizebox{\columnwidth}{!}{%
    \begin{tabular}{crrrr}
    \toprule
    Dataset  & \#Manip & \#Non-manip & \#Tech & \#Vul \\
    \toprule
    \datasetnamecon  & $2,016$ & $~~~~899$ & $1,748$ & $605$ \\
    \midrule
    \datasetnamemaj & $2,818$ & $1,182$ & $2,154$ & $731$ \\
    \bottomrule
    \end{tabular}
    }
    \vspace{-1mm}
    \caption{Number of manipulative and non-manipulative dialogues, and manipulative dialogues that contain technique and vulnerability elements.}
    \vspace{-4mm}
    \label{tab:final_label_2}
\end{table}

%% file: tables/gender_agreement.tex
\begin{table}[t]
    \centering
    \small
    \resizebox{\columnwidth}{!}{%
    \begin{tabular}{cccc}
    \toprule
    Gender (Count)  & Female~($14$) & Male~($3$) & Avg. Conf. \\
    \toprule
    Female~($14$) & $0.82$ & $0.83$ & $3.61$ \\
    \midrule
    Male~($3$) & $0.83$ & $0.91$ & $3.96$ \\
    \bottomrule
    \end{tabular}
    }
    \caption{\textcolor{\highlightcolor}{The average inter-annotator agreement scores across female and male annotators, and the average confidence scores of female and male annotators.}}
    \vspace{-2mm}
    \label{tab:gender_agreement}
\end{table}

%% file: tables/annotation_sample.tex
\begin{table*}[t!]
\setlength{\tabcolsep}{2.2pt}
\centering 
\resizebox{\textwidth}{!}{%
\begin{tabular}{p{10cm}ccccc} 
\toprule
\multirow{2}{*}{Dialogue} & \multicolumn{5}{c}{Annotation Example} \\
\cmidrule{2-6}
& Manipulative & Technique & Victim & Vulnerability  & Confidence \\ 
\toprule
Person1: \colorbox{green}{I like you so much. I think you're beautiful. I think if} \colorbox{green}{we were together you would love it. You wouldn't believe it.} \newline 
Person2: How do you know? \newline 
Person1: \colorbox{green}{I just know. I know you'll love it.} \newline
Person2: But I'm scared Telly. \newline
Person1: I'm telling you. There's nothing in the world to worry about. \newline
Person2: Nothing? & Yes & Persuasion or Seduction & Yes & Naivete & 4 \\
\bottomrule
\end{tabular}
}
\vspace{-1mm}
\caption{An example of annotation. Highlighted text indicates parts identified as manipulative by the annotator.}
\vspace{-2mm}
\label{tab:annotation_example} 
\end{table*}

%% file: annotation_observation.tex
\section{Annotator Feedback}
\label{appendix:feedback}
\subsection{Prior Knowledge of Dialogue}

There were several observations derived from the experiences of the annotators. Firstly, there was the incidence of prior knowledge of dialogue. The dataset was derived from dialogue in movie scripts, and as such, did include recognizable dialogue from some more well-known movie titles, such as ``The Talented Mr. Ripley''. Given that annotators had more background knowledge with regards to the dialogue, and greater context, there is possibility that their annotation choices could have been influenced by their prior exposure to and knowledge of the movie dialogue. 

\subsection{Mutual Manipulation}

Another observation from the annotation experience was that there could be mutual manipulation weaponized by both parties within a dialogue. While some dialogue clearly reflected manipulative speech by one party on the other, certain dialogues showcased manipulative tactics on both sides. Thus, it becomes difficult to differentiate between a clear perpetrator and victim, which also influences the selection of manipulation techniques during the annotation process.

\subsection{Cognitive Fatigue / Overanalysis}

Lastly, annotators reported cognitive fatigue and over-analysis of tasks. Throughout the annotation process, which usually consisted of individual extended sessions of annotating, individuals became overly sensitive to cues and patterns. This hypersensitivity led to a heightened perception of manipulation in dialogues, such that they were unable to maintain a balanced perspective.

%% file: main.bbl
\begin{thebibliography}{38}
\expandafter\ifx\csname natexlab\endcsname\relax\def\natexlab#1{#1}\fi

\bibitem[{Baheti et~al.(2021)Baheti, Sap, Ritter, and Riedl}]{bahetiJustSayNo2021}
Ashutosh Baheti, Maarten Sap, Alan Ritter, and Mark Riedl. 2021.
\newblock \href {https://doi.org/10.18653/v1/2021.emnlp-main.397} {Just say no: Analyzing the stance of neural dialogue generation in offensive contexts}.
\newblock In \emph{Proceedings of the 2021 Conference on Empirical Methods in Natural Language Processing}, pages 4846--4862. Association for Computational Linguistics.

\bibitem[{Barnhill(2014)}]{barnhillManipulation2014}
Anne Barnhill. 2014.
\newblock \href {https://doi.org/10.1093/acprof:oso/9780199338207.003.0003} {What is manipulation?}
\newblock In \emph{Manipulation: Theory and Practice}. Oxford University Press.

\bibitem[{Barnhill(2022)}]{barnhill2022philosophy}
Anne Barnhill. 2022.
\newblock How philosophy might contribute to the practical ethics of online manipulation.
\newblock In \emph{The philosophy of online manipulation}, pages 49--71. Routledge.

\bibitem[{Basile et~al.(2019)Basile, Bosco, Fersini, Nozza, Patti, Rangel~Pardo, Rosso, and Sanguinetti}]{basileSemEval2019TaskMultilingual2019}
Valerio Basile, Cristina Bosco, Elisabetta Fersini, Debora Nozza, Viviana Patti, Francisco~Manuel Rangel~Pardo, Paolo Rosso, and Manuela Sanguinetti. 2019.
\newblock \href {https://doi.org/10.18653/v1/S19-2007} {{S}em{E}val-2019 task 5: Multilingual detection of hate speech against immigrants and women in {T}witter}.
\newblock In \emph{Proceedings of the 13th International Workshop on Semantic Evaluation}, pages 54--63, {Minneapolis, Minnesota, USA}. {Association for Computational Linguistics}.

\bibitem[{Bubeck et~al.(2023)Bubeck, Chandrasekaran, Eldan, Gehrke, Horvitz, Kamar, Lee, Lee, Li, Lundberg, Nori, Palangi, Ribeiro, and Zhang}]{bubeckSparksArtificialGeneral2023}
S{\'e}bastien Bubeck, Varun Chandrasekaran, Ronen Eldan, Johannes Gehrke, Eric Horvitz, Ece Kamar, Peter Lee, Yin~Tat Lee, Yuanzhi Li, Scott Lundberg, Harsha Nori, Hamid Palangi, Marco~Tulio Ribeiro, and Yi~Zhang. 2023.
\newblock \href {https://doi.org/10.48550/arXiv.2303.12712} {Sparks of artificial general intelligence: Early experiments with gpt-4}.
\newblock \emph{arXiv preprint arXiv:2303.12712}.

\bibitem[{Coppersmith et~al.(2018)Coppersmith, Leary, Crutchley, and Fine}]{coppersmithNaturalLanguageProcessing2018}
Glen Coppersmith, Ryan Leary, Patrick Crutchley, and Alex Fine. 2018.
\newblock \href {https://doi.org/10.1177/1178222618792860} {Natural language processing of social media as screening for suicide risk}.
\newblock \emph{Biomedical Informatics Insights}, 10:1178222618792860.

\bibitem[{Danescu-Niculescu-Mizil and Lee(2011)}]{danescu-niculescu-mizilChameleonsImaginedConversations2011}
Cristian Danescu-Niculescu-Mizil and Lillian Lee. 2011.
\newblock \href {https://aclanthology.org/W11-0609} {Chameleons in imagined conversations: A new approach to understanding coordination of linguistic style in dialogs}.
\newblock In \emph{Proceedings of the 2nd Workshop on Cognitive Modeling and Computational Linguistics}, pages 76--87, Portland, Oregon, USA. {Association for Computational Linguistics}.

\bibitem[{De~Choudhury et~al.(2016)De~Choudhury, Kiciman, Dredze, Coppersmith, and Kumar}]{dechoudhuryDiscoveringShiftsSuicidal2016}
Munmun De~Choudhury, Emre Kiciman, Mark Dredze, Glen Coppersmith, and Mrinal Kumar. 2016.
\newblock \href {https://doi.org/10.1145/2858036.2858207} {Discovering shifts to suicidal ideation from mental health content in social media}.
\newblock In \emph{Proceedings of the SIGCHI conference on human factors in computing systems. CHI Conference}, volume 2016, pages 2098--2110.

\bibitem[{Deng et~al.(2023)Deng, Cheng, Sun, Zhang, and Huang}]{dengRecentAdvancesSafe2023}
Jiawen Deng, Jiale Cheng, Hao Sun, Zhexin Zhang, and Minlie Huang. 2023.
\newblock \href {http://arxiv.org/abs/2302.09270} {Towards safer generative language models: A survey on safety risks, evaluations, and improvements}.

\bibitem[{Dixon et~al.(2018)Dixon, Li, Sorensen, Thain, and Vasserman}]{dixonMeasuringMitigatingUnintended2018}
Lucas Dixon, John Li, Jeffrey Sorensen, Nithum Thain, and Lucy Vasserman. 2018.
\newblock \href {https://doi.org/10.1145/3278721.3278729} {Measuring and mitigating unintended bias in text classification}.
\newblock In \emph{Proceedings of the 2018 AAAI/ACM Conference on AI, Ethics, and Society}, pages 67--73, New Orleans LA USA. {ACM}.

\bibitem[{Eichstaedt et~al.(2018)Eichstaedt, Smith, Merchant, Ungar, Crutchley, {Preo{\c t}iuc-Pietro}, Asch, and Schwartz}]{eichstaedtFacebookLanguagePredicts2018}
Johannes~C. Eichstaedt, Robert~J. Smith, Raina~M. Merchant, Lyle~H. Ungar, Patrick Crutchley, Daniel {Preo{\c t}iuc-Pietro}, David~A. Asch, and H.~Andrew Schwartz. 2018.
\newblock \href {https://doi.org/10.1073/pnas.1802331115} {Facebook language predicts depression in medical records}.
\newblock In \emph{Proceedings of the National Academy of Sciences}, volume 115, pages 11203--11208. {Proceedings of the National Academy of Sciences}.

\bibitem[{Fleiss and Cohen(1973)}]{fleissEquivalenceWeightedKappa1973}
Joseph~L. Fleiss and Jacob Cohen. 1973.
\newblock \href {https://doi.org/10.1177/001316447303300309} {The equivalence of weighted kappa and the intraclass correlation coefficient as measures of reliability}.
\newblock \emph{Educational and Psychological Measurement}, 33:613--619.

\bibitem[{Gao and Huang(2017)}]{gaoDetectingOnlineHate2017}
Lei Gao and Ruihong Huang. 2017.
\newblock \href {https://doi.org/10.26615/978-954-452-049-6_036} {Detecting online hate speech using context aware models}.
\newblock In \emph{Proceedings of the International Conference Recent Advances in Natural Language Processing}, pages 260--266. Incoma Ltd. Shoumen, Bulgaria.

\bibitem[{Guntuku et~al.(2019)Guntuku, Buffone, Jaidka, Eichstaedt, and Ungar}]{guntukuUnderstandingMeasuringPsychological2019}
Sharath~Chandra Guntuku, Anneke Buffone, Kokil Jaidka, Johannes~C. Eichstaedt, and Lyle~H. Ungar. 2019.
\newblock \href {https://doi.org/10.1609/icwsm.v13i01.3223} {Understanding and measuring psychological stress using social media}.
\newblock In \emph{Proceedings of the International AAAI Conference on Web and Social Media}, volume~13, pages 214--225.

\bibitem[{Hamel et~al.(2023)Hamel, Cannon, and {Graham-Kevan}}]{hamelAbuse2023}
John Hamel, Clare E.~B. Cannon, and Nicola {Graham-Kevan}. 2023.
\newblock \href {https://doi.org/10.1037/trm0000449} {The consequences of psychological abuse and control in intimate partner relationships.}
\newblock \emph{Traumatology}.

\bibitem[{Haque et~al.(2021)Haque, Reddi, and Giallanza}]{haqueDeepLearningSuicide2021}
Ayaan Haque, Viraaj Reddi, and Tyler Giallanza. 2021.
\newblock \href {https://doi.org/10.1007/978-3-030-86383-8_35} {Deep learning for suicide and depression identification with unsupervised label correction}.
\newblock In \emph{Artificial Neural Networks and Machine Learning}, pages 436--447. Springer International Publishing.

\bibitem[{Hartvigsen et~al.(2022)Hartvigsen, Gabriel, Palangi, Sap, Ray, and Kamar}]{hartvigsenToxiGen2022}
Thomas Hartvigsen, Saadia Gabriel, Hamid Palangi, Maarten Sap, Dipankar Ray, and Ece Kamar. 2022.
\newblock \href {https://doi.org/10.18653/v1/2022.acl-long.234} {{T}oxi{G}en: A large-scale machine-generated dataset for adversarial and implicit hate speech detection}.
\newblock In \emph{Proceedings of the 60th Annual Meeting of the Association for Computational Linguistics}, pages 3309--3326, Dublin, Ireland. Association for Computational Linguistics.

\bibitem[{Hosseinmardi et~al.(2015)Hosseinmardi, Mattson, Rafiq, Han, Lv, and Mishra}]{hosseinmardiDetectionCyberbullyingIncidents2015}
Homa Hosseinmardi, Sabrina~Arredondo Mattson, Rahat~Ibn Rafiq, Richard Han, Qin Lv, and Shivakant Mishra. 2015.
\newblock \href {http://arxiv.org/abs/1503.03909} {Detection of cyberbullying incidents on the instagram social network}.
\newblock \emph{arXiv preprint arXiv:1503.03909}.

\bibitem[{Ienca(2023)}]{iencaAIManipulation2023}
Marcello Ienca. 2023.
\newblock \href {https://doi.org/10.1007/s11245-023-09940-3} {On artificial intelligence and manipulation}.
\newblock \emph{Topoi}, 42(3):833--842.

\bibitem[{Liu et~al.(2019)Liu, Ott, Goyal, Du, Joshi, Chen, Levy, Lewis, Zettlemoyer, and Stoyanov}]{liuRoBERTaRobustlyOptimized2019}
Yinhan Liu, Myle Ott, Naman Goyal, Jingfei Du, Mandar Joshi, Danqi Chen, Omer Levy, Mike Lewis, Luke Zettlemoyer, and Veselin Stoyanov. 2019.
\newblock \href {https://doi.org/10.48550/arXiv.1907.11692} {Roberta: A robustly optimized bert pretraining approach}.
\newblock \emph{arXiv preprint arXiv:1907.11692}.

\bibitem[{Miao et~al.(2020)Miao, Last, and Litvak}]{miaoDetectingTrollTweets2020}
Lin Miao, Mark Last, and Marina Litvak. 2020.
\newblock \href {https://aclanthology.org/2020.lrec-1.766} {Detecting troll tweets in a bilingual corpus}.
\newblock In \emph{Proceedings of the Twelfth Language Resources and Evaluation Conference}, pages 6247--6254, Marseille, France. European Language Resources Association.

\bibitem[{Mishra et~al.(2020)Mishra, Yannakoudakis, and Shutova}]{mishraTacklingOnlineAbuse2020}
Pushkar Mishra, Helen Yannakoudakis, and Ekaterina Shutova. 2020.
\newblock \href {http://arxiv.org/abs/1908.06024} {Tackling online abuse: A survey of automated abuse detection methods}.
\newblock \emph{arXiv preprint arXiv:1908.06024}.

\bibitem[{Naseem et~al.(2022)Naseem, Dunn, Kim, and Khushi}]{naseemEarlyIdentificationDepression2022}
Usman Naseem, Adam~G. Dunn, Jinman Kim, and Matloob Khushi. 2022.
\newblock \href {https://doi.org/10.1145/3485447.3512128} {Early {{Identification}} of {{Depression Severity Levels}} on {{Reddit Using Ordinal Classification}}}.
\newblock In \emph{Proceedings of the {{ACM Web Conference}} 2022}, {{WWW}} '22, pages 2563--2572, {New York, NY, USA}. {Association for Computing Machinery}.

\bibitem[{Nijhawan et~al.(2022)Nijhawan, Attigeri, and Ananthakrishna}]{nijhawanStressDetectionUsing2022}
Tanya Nijhawan, Girija Attigeri, and T.~Ananthakrishna. 2022.
\newblock \href {https://doi.org/10.1186/s40537-022-00575-6} {Stress detection using natural language processing and machine learning over social interactions}.
\newblock \emph{Journal of Big Data}, 9(1):33.

\bibitem[{Rosa et~al.(2018)Rosa, Carvalho, Calado, Martins, Ribeiro, and Coheur}]{rosaUsingFuzzyFingerprints2018}
Hugo Rosa, Joao~P. Carvalho, P{\'a}vel Calado, Bruno Martins, Ricardo Ribeiro, and Luisa Coheur. 2018.
\newblock \href {https://doi.org/10.1109/FUZZ-IEEE.2018.8491557} {Using fuzzy fingerprints for cyberbullying detection in social networks}.
\newblock In \emph{2018 IEEE International Conference on Fuzzy Systems}, pages 1--7, Rio de Janeiro, Brazil. IEEE Press.

\bibitem[{Simon and Foley(2011)}]{simon2011sheep}
George~K Simon and Kevin Foley. 2011.
\newblock \emph{In sheep's clothing: Understanding and dealing with manipulative people}.
\newblock Tantor Media, Incorporated.

\bibitem[{Touvron et~al.(2023)Touvron, Lavril, Izacard, Martinet, Lachaux, Lacroix, Rozi{\`e}re, Goyal, Hambro, Azhar, Rodriguez, Joulin, Grave, and Lample}]{touvronLLaMAOpenEfficient2023}
Hugo Touvron, Thibaut Lavril, Gautier Izacard, Xavier Martinet, Marie-Anne Lachaux, Timoth{\'e}e Lacroix, Baptiste Rozi{\`e}re, Naman Goyal, Eric Hambro, Faisal Azhar, Aurelien Rodriguez, Armand Joulin, Edouard Grave, and Guillaume Lample. 2023.
\newblock \href {https://doi.org/10.48550/arXiv.2302.13971} {Llama: Open and efficient foundation language models}.
\newblock \emph{arXiv preprint arXiv:2302.13971}.

\bibitem[{Turcan and McKeown(2019)}]{turcanDreadditRedditDataset2019}
Elsbeth Turcan and Kathy McKeown. 2019.
\newblock \href {https://doi.org/10.18653/v1/D19-6213} {Dreaddit: A reddit dataset for stress analysis in social media}.
\newblock In \emph{Proceedings of the Tenth International Workshop on Health Text Mining and Information Analysis}, pages 97--107, {Hong Kong}. {Association for Computational Linguistics}.

\bibitem[{Wang et~al.(2023)Wang, Chen, Pei, Xie, Kang, Zhang, Xu, Xiong, Dutta, Schaeffer, Truong, Arora, Mazeika, Hendrycks, Lin, Cheng, Koyejo, Song, and Li}]{wangDecodingTrustComprehensiveAssessment2023}
Boxin Wang, Weixin Chen, Hengzhi Pei, Chulin Xie, Mintong Kang, Chenhui Zhang, Chejian Xu, Zidi Xiong, Ritik Dutta, Rylan Schaeffer, Sang~T. Truong, Simran Arora, Mantas Mazeika, Dan Hendrycks, Zinan Lin, Yu~Cheng, Sanmi Koyejo, Dawn Song, and Bo~Li. 2023.
\newblock \href {https://openreview.net/forum?id=kaHpo8OZw2} {Decodingtrust: A comprehensive assessment of trustworthiness in {GPT} models}.
\newblock In \emph{Thirty-seventh Conference on Neural Information Processing Systems Datasets and Benchmarks Track}.

\bibitem[{Wang and Potts(2019)}]{wangTalkDownCorpusCondescension2019}
Zijian Wang and Christopher Potts. 2019.
\newblock \href {https://doi.org/10.18653/v1/D19-1385} {Talkdown: A corpus for condescension detection in context}.
\newblock In \emph{Proceedings of the 2019 Conference on Empirical Methods in Natural Language Processing and the 9th International Joint Conference on Natural Language Processing}, pages 3709--3717, Hong Kong, China. Association for Computational Linguistics.

\bibitem[{Waseem(2016)}]{waseemAreYouRacist2016}
Zeerak Waseem. 2016.
\newblock \href {https://doi.org/10.18653/v1/W16-5618} {Are you a racist or am i seeing things? annotator influence on hate speech detection on twitter}.
\newblock In \emph{Proceedings of the First Workshop on NLP and Computational Social Science}, pages 138--142, {Austin, Texas}. {Association for Computational Linguistics}.

\bibitem[{Wiegand et~al.(2021)Wiegand, Ruppenhofer, and Eder}]{wiegandImplicitlyAbusiveLanguage2021}
Michael Wiegand, Josef Ruppenhofer, and Elisabeth Eder. 2021.
\newblock \href {https://doi.org/10.18653/v1/2021.naacl-main.48} {Implicitly abusive language {--} what does it actually look like and why are we not getting there?}
\newblock In \emph{Proceedings of the 2021 Conference of the North American Chapter of the Association for Computational Linguistics: Human Language Technologies}, pages 576--587, Online. Association for Computational Linguistics.

\bibitem[{Xu et~al.(2019)Xu, Chikersal, Doryab, Villalba, Dutcher, Tumminia, Althoff, Cohen, Creswell, Creswell, Mankoff, and Dey}]{xuLeveragingRoutineBehavior2019}
Xuhai Xu, Prerna Chikersal, Afsaneh Doryab, Daniella~K. Villalba, Janine~M. Dutcher, Michael~J. Tumminia, Tim Althoff, Sheldon Cohen, Kasey~G. Creswell, J.~David Creswell, Jennifer Mankoff, and Anind~K. Dey. 2019.
\newblock \href {https://doi.org/10.1145/3351274} {Leveraging {{Routine Behavior}} and {{Contextually-Filtered Features}} for {{Depression Detection}} among {{College Students}}}.
\newblock In \emph{Proceedings of the ACM on Interactive, Mobile, Wearable and Ubiquitous Technologies}, volume~3, pages 116:1--116:33.

\bibitem[{Xu et~al.(2023)Xu, Yao, Dong, Gabriel, Yu, Hendler, Ghassemi, Dey, and Wang}]{xuMentalLLMLeveragingLarge2023}
Xuhai Xu, Bingsheng Yao, Yuanzhe Dong, Saadia Gabriel, Hong Yu, James Hendler, Marzyeh Ghassemi, Anind~K. Dey, and Dakuo Wang. 2023.
\newblock Mental-llm: Leveraging large language models for mental health prediction via online text data.
\newblock \emph{arXiv preprint arXiv:2307.14385}.

\bibitem[{Yang et~al.(2023)Yang, Ji, Zhang, Xie, Kuang, and Ananiadou}]{yangInterpretableMentalHealth2023}
Kailai Yang, Shaoxiong Ji, Tianlin Zhang, Qianqian Xie, Ziyan Kuang, and Sophia Ananiadou. 2023.
\newblock \href {https://doi.org/10.18653/v1/2023.emnlp-main.370} {Towards interpretable mental health analysis with large language models}.
\newblock In \emph{Proceedings of the 2023 Conference on Empirical Methods in Natural Language Processing}, pages 6056--6077, Singapore. Association for Computational Linguistics.

\bibitem[{Yavnyi et~al.(2023)Yavnyi, Sliusarenko, Razzaghi, Nahorna, Mo, Hovakimyan, and Chernodub}]{yavnyiDeTexDBenchmarkDataset2023}
Serhii Yavnyi, Oleksii Sliusarenko, Jade Razzaghi, Olena Nahorna, Yichen Mo, Knar Hovakimyan, and Artem Chernodub. 2023.
\newblock \href {https://doi.org/10.18653/v1/2023.woah-1.2} {{{DeTexD}}: {{A Benchmark Dataset}} for {{Delicate Text Detection}}}.
\newblock In \emph{The 7th {{Workshop}} on {{Online Abuse}} and {{Harms}} ({{WOAH}})}, pages 14--28, {Toronto, Canada}. {Association for Computational Linguistics}.

\bibitem[{Yin and Zubiaga(2022)}]{yinHiddenObviousMisleading2022}
Wenjie Yin and Arkaitz Zubiaga. 2022.
\newblock \href {https://doi.org/10.1016/j.osnem.2022.100210} {Hidden behind the obvious: Misleading keywords and implicitly abusive language on social media}.
\newblock \emph{Online Social Networks and Media}, 30:100210.

\bibitem[{Zhang et~al.(2021)Zhang, Ren, and De~Rijke}]{zhangTaxonomyDataSet2021}
Yangjun Zhang, Pengjie Ren, and Maarten De~Rijke. 2021.
\newblock \href {https://doi.org/10.1002/asi.24496} {A taxonomy, data set, and benchmark for detecting and classifying malevolent dialogue responses}.
\newblock \emph{Journal of the Association for Information Science and Technology}, 72(12):1477--1497.

\end{thebibliography}
